\documentclass[10pt,journal,compsoc]{IEEEtran}

\usepackage{textcomp}
\usepackage{stfloats}
\usepackage{url}
\usepackage{verbatim}
\usepackage{graphicx}
\usepackage{stfloats}
\usepackage{amsfonts}
\usepackage{makecell}
\usepackage{pifont} 
\usepackage{amsmath}
\usepackage{booktabs}
\usepackage{ragged2e}
\usepackage[table]{xcolor}
\usepackage[breaklinks=true,bookmarks=false,urlcolor=black,colorlinks,citecolor=green]{hyperref}

\newcommand{\HF}[1]{\textcolor[rgb]{1.00,0.00,0.00}{#1}}

\newcommand{\red}[1]{\textcolor[rgb]{1.00,0.00,0.00}{#1}}

\newcommand{\eg}{\emph{e.g.}}
\newcommand{\ie}{\emph{i.e.}}

\ifCLASSOPTIONcompsoc
  \usepackage[nocompress]{cite}
\else
  \usepackage{cite}
\fi

\ifCLASSINFOpdf
  
\else
  
\fi

\begin{document}
%
\title{Robust Domain Adaptive Object Detection with \\ Unified Multi-Granularity Alignment}
%
%
%
%

\author{Libo~Zhang, Wenzhang Zhou, Heng Fan, Tiejian~Luo, and Haibin Ling

\IEEEcompsocitemizethanks{

\IEEEcompsocthanksitem Libo Zhang is with the State Key Laboratory of Computer Science, Institute of Software Chinese Academy of Sciences, China. E-mail: libo@iscas.ac.cn.

\IEEEcompsocthanksitem Wenzhang Zhou and Tiejian Luo are with the University of Chinese Academy of Sciences, China. E-mail: zhouwenzhang19@mails.ucas.ac.cn, tjluo@ucas.ac.cn.

\IEEEcompsocthanksitem Heng Fan is with the Department of Computer Science and Engineering, University of North Texas, USA. E-mail: heng.fan@unt.edu.

\IEEEcompsocthanksitem Haibin Ling is with the Department of Computer Science, Stony Brook University, USA. E-mail: hling@cs.stonybrook.edu. 

\IEEEcompsocthanksitem Libo Zhang and Wenzhang Zhou make equal contributions to this work. 

\IEEEcompsocthanksitem Heng Fan is the corresponding author.

\IEEEcompsocthanksitem A preliminary version~\cite{zhou2022multi} of this work has appeared in CVPR 2022.

}


}

%
%

\markboth{Submitted to IEEE Journal}%
{Shell \MakeLowercase{\textit{et al.}}: Bare Demo of IEEEtran.cls for Computer Society Journals}

\IEEEtitleabstractindextext{%
\begin{abstract}

\justifying Domain adaptive detection aims to improve the generalization of detectors on target domain. To reduce discrepancy in feature distributions between two domains, recent approaches achieve domain adaption through feature alignment in different granularities via adversarial learning. However, they neglect the relationship between multiple granularities and different features in alignment, degrading detection. Addressing this, we introduce a unified multi-granularity alignment (MGA)-based detection framework for domain-invariant feature learning. The key is to encode the dependencies across different granularities including pixel-, instance-, and category-levels simultaneously to align two domains. Specifically, based on pixel-level features, we first develop an omni-scale gated fusion (OSGF) module to aggregate discriminative representations of instances with scale-aware convolutions, leading to robust multi-scale detection. Besides, we introduce multi-granularity discriminators to identify where, either source or target domains, different granularities of samples come from. Note that, MGA not only leverages instance discriminability in different categories but also exploits category consistency between two domains for detection. Furthermore, we present an adaptive exponential moving average (AEMA) strategy that explores model assessments for model update to improve pseudo labels and alleviate local misalignment problem, boosting detection robustness. Extensive experiments on multiple domain adaption scenarios validate the superiority of MGA over other approaches on FCOS and Faster R-CNN detectors. Code will be released at \url{https://github.com/tiankongzhang/MGA}.
\end{abstract}

}

\maketitle

\IEEEdisplaynontitleabstractindextext

\IEEEpeerreviewmaketitle

\IEEEraisesectionheading{\section{Introduction}\label{sec:introduction}}

%
\IEEEPARstart{O}{bject} detection has been one of the most fundamental problems in computer vision with a long list of applications such as visual surveillance, self-driving, robotics, etc. Owing to the powerful representation by deep learning (\eg,~\cite{DBLP:conf/cvpr/HeZRS16,krizhevsky2012imagenet,DBLP:journals/corr/SimonyanZ14a}), object detection has witnessed considerable advancement in recent years with numerous excellent frameworks (\eg,~\cite{girshick2014rich,girshick2015fast,DBLP:journals/pami/RenHG017,DBLP:conf/iccv/TianSCH19,DBLP:conf/eccv/LawD18,liu2016ssd,redmon2016you,lin2017feature,liang2023clusterfomer}). These modern detectors are usually trained and evaluated on a large-scale annotated dataset (\eg,~\cite{lin2014microsoft}). Despite the great achievement, they may suffer from poor generalization when applied to images from a new target domain. To remedy this, a simple and straightforward solution is to build a benchmark for the new target domain and re-train the detector. Nevertheless, benchmark creation is both time-consuming and costly. In addition, the new target domain could arbitrary and it is almost \textit{impossible} to develop benchmarks for all new target domains. 

\begin{figure}[!t]
\centering
\includegraphics[width=\linewidth]{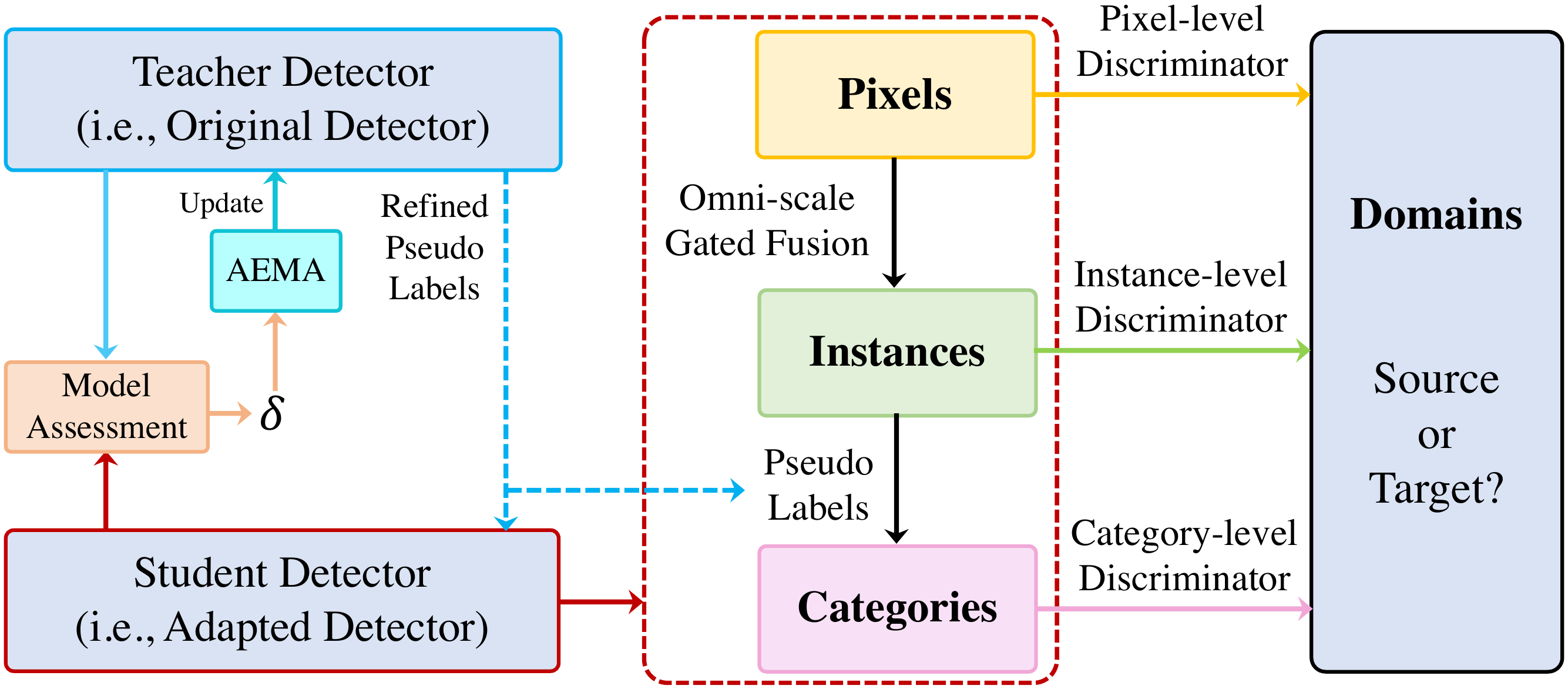}
\caption{Illustration of the proposed Multi-Granularity Alignment (\textbf{MGA}) framework for domain adaptive object detection. Specifically, MGA encodes the dependencies across multiple granularities simultaneously, including pixel-, instance-, and category-levels. In addition, a dynamic update mechanism guided by update factor $\delta$ (as detailed later) through model assessment during training is used to improve the quality of pseudo labels and meanwhile mitigate the local misalignment problem, further enhancing the detection robustness. Best viewed in color and by zooming in for all figures throughout this paper.}
\label{fig:fig1}
\end{figure}

In order to deal with the above issues, researchers have explored the \textit{unsupervised domain adaption} (UDA) detection, with the goal of transferring knowledge learned from an annotated source domain to an unlabeled target domain. One popular trend is to leverage adversarial learning~\cite{DBLP:conf/icml/GaninL15} to narrow the discrepancy between domains. Specifically, a domain discriminator is introduced to distinguish which domain, the source or the target, the image comes from. Then, the detector learns the domain-invariant feature representation by confusing the discriminator~\cite{DBLP:conf/cvpr/SaitoUHS19}. Despite achieving promising results, previous domain adaption approaches may suffer from target scale variations in cluttered regions due to the fixed kernel design in ConvNets~\cite{DBLP:conf/cvpr/HeZRS16,krizhevsky2012imagenet}, which results in difficulty in learning discriminative representations for objects of different scales and thus degrades detection performance. For example, the features of small targets may contain too much background noise because of too large receptive field in the convolutional layer. Meanwhile, the features of large objects may lack global structural information owing to too small receptive field. In addition, the intrinsic relation of feature distributions between two domains are neglected.

To address the aforementioned problems, various feature alignment strategies have been introduced in the adversarial learning manner~\cite{DBLP:conf/icml/GaninL15,tzeng2017adversarial} for better target domain adaption. These alignment approaches can be summarized into three categories based on different granularity perspectives, consisting of pixel-, instance-, and category-level. \textit{Pixel-level alignment}~\cite{DBLP:conf/cvpr/KimJKCK19,DBLP:conf/eccv/HsuTLY20,hsu2020progressive} aims at aligning lower-level pixel feature distribution of objects and background regions. Nevertheless, there may exist a large gap between the pixel-level features for objects of different scales within the same category, resulting in limited detection performance. Different from pixel-level alignment, \textit{instance-level alignment}~\cite{cai2019exploring,he2019multi,DBLP:conf/cvpr/Chen0SDG18,DBLP:conf/eccv/LiDZWLWZ20,zhu2019adapting} first pools the feature maps of detection proposals and then leverages the pooled proposal features for the domain discriminator training. Despite avoiding the gap in pixel-level alignment, this strategy suffers from feature distortion for objects of different scales and aspect ratios caused by pooling operation, which may lead to inaccurate feature representation and degenerated results. Besides the aforementioned two types of alignments, recent approaches have attempted to utilize \textit{category-level alignment}~\cite{DBLP:conf/iccv/DuTYFXZYZ19, DBLP:conf/cvpr/HuKSC20, DBLP:conf/eccv/PaulTSRC20, DBLP:conf/eccv/WangSZD020, DBLP:conf/cvpr/XuZJW20, DBLP:conf/cvpr/XuWNTZ20} for UDA detection. In specific, considering the intrinsic relation of feature distributions in two domains, the category-level alignment applies the categorical discriminability to handle hard aligned instances. However, this alignment mechanism focuses only on the global consistency of feature distribution between two domains while ignores other local consistency constrains.

\vspace{0.5em}

\noindent
\textbf{Contribution.} Although each type of alignment strategy brings in improvement, they are limited in several aspects as discussed above. In order to address these issues and make full use of the advantages of these three alignments, we propose a novel unified Multi-Granularity Alignment (MGA) framework for UDA detection, as shown in Figure~\ref{fig:fig1}.

Instead of performing simple combination of different alignment methods, MGA simultaneously encodes the dependencies across different granularities, consisting of pixel-, instance-, and category-levels, for domain alignment. More specifically, we first introduce an omni-scale gated fusion (OSGF) module in MGA. The OSGF module is able to adapt to instances of different scales by automatically choosing the most plausible convolutions from the
low- and high-resolution streams for feature extraction. Concretely, we first predict coarse detections based on pixel-level backbone feature maps. Then, using these coarse detections as guidance, we design a set of parallel convolutions in OSGF and adopt a gate mechanism to aggregate the discriminative features of instances (\ie, the coarse detections) with similar scales and aspect ratios. By doing so, the following detection head can more accurately predict multi-scale objects. 

Besides the OSGF module, we present a new category-level discriminator. Different from previous approaches, our category-level discriminator takes into account not only the instance discriminability in different classes but also the category consistency between two domains, leading to better detection. In order to supervise the category-level discriminator, pseudo labels are assigned to important instances with high confidence based on object detection results. 

Considering that the quality of pseudo labels is crucial for learning a good category-level discriminator, we propose a simple yet effective adaptive exponential moving average (AEMA) strategy to train the teacher detector (\ie, the original detector). As shown in Figure~\ref{fig:fig1}, during the training phase, we assess both the teacher detector and the student detector (\ie, the final adaptive detector) on the source domain. Based on their model assessments, a dynamic update factor $\delta$ (as described later) is learned and utilized as a guidance to adjust the coefficient parameter in exponential moving average (EMA) update in an adaptive manner. The resulted AEMA helps better train the teacher detector to produce high-quality pseudo labels and meanwhile alleviate the local misalignment caused by low-quality pseudo labels, significantly enhancing the detection robustness. We will elaborate on the details of our AEMA later.

By developing the multi-granularity discriminators, our MGA exploits and integrates rich complementary information from different levels, and hence achieves better UDA detection. Besides, the proposed AEMA further enhances the robustness with high-quality pseudo labels. 

To validate the effectiveness of our approach, we carry out extensive experiments on multiple domain-shift scenarios using various benchmarks including Cityscapes~\cite{DBLP:conf/cvpr/CordtsORREBFRS16}, FoggyCityscapes~\cite{DBLP:journals/ijcv/SakaridisDG18}, Sim10k~\cite{DBLP:/conf/icra/driving17}, KITTI~\cite{DBLP:/conf/cvpr/are12}, PASCAL VOC~\cite{DBLP:journals/ijcv/EveringhamGWWZ10}, Clipart~\cite{DBLP:conf/cvpr/InoueFYA18} and Watercolor~\cite{DBLP:conf/cvpr/InoueFYA18}. We evaluate the proposed method on the top of two popular detection frameworks, the anchor-free FCOS~\cite{DBLP:conf/iccv/TianSCH19} and the anchor-based Faster R-CNN~\cite{DBLP:journals/pami/RenHG017}, with VGG-16~\cite{DBLP:journals/corr/SimonyanZ14a} and ResNet-101~\cite{DBLP:conf/cvpr/HeZRS16} backbones. Experiment results demonstrate that our MGA together with the dynamic model update significantly improve the baseline detectors with superior results over other state-of-the-arts.

To sum up, we make the following key contributions:

\begin{itemize}
    \item We propose a novel unified multi-granularity alignment (MGA) framework that encodes dependencies across different pixel-, instance-, and category-levels for UDA detection. Notably, our MGA framework is general and applicable to different object detectors.

    \item We present an omni-scale gate fusion (OSGF) module to extract discriminative feature representation for instances with different scales and aspect ratios.

    \item we propose a new category-level discriminator by exploiting both the instance discriminability in different classes and the category consistency between two domains, leading to better detection.

    \item We introduce a simple yet effective dynamic adaptive exponential moving average (AEMA) strategy to improve the quality of pseudo labels and meanwhile mitigate the local misalignment issue in UDA detection, which significantly boosts the robustness of detector.

    \item On extensive experiments on multiple domain adaption scenarios, the proposed approach outperforms other state-of-the-art UDA detectors on the top of two detection frameworks, evidencing its effectiveness and generality.
\end{itemize}

This paper builds upon our preliminary conference version~\cite{zhou2022multi} and significantly extends it in different aspects. \textbf{(1)} We propose an effective assessment-based AEMA for model update of the teacher detector. This way, we are able to obtain pseudo labels with better quality, which largely boosts the detection robustness. Meanwhile, it is beneficial in alleviating the local misalignment issue caused by low-quality pseudo labels, further improving the performance. \textbf{(2)} We modify the structure of the OSGF module by sharing the convolutional layer (as described later). Note that, this modification is not trivial. It not only brings in improvement on the detection results but also decreases the number of parameters. \textbf{(3)} We incorporate more experiments and comparisons with in-depth analysis and ablation studies to further show the effectiveness of our approach. \textbf{(4)} We supplement thorough visual analysis of our detector, which allows the readers to better understand our method.

The rest of this paper is organized as follows. Section~\ref{rela} discusses approaches related to this paper. Our approach is elaborated in Section~\ref{mga}. In Section~\ref{exp}, we demonstrate the experimental results, including comparisons with state-of-the-arts, ablation studies and visual analysis. Section~\ref{diss} presents discussions, followed by conclusion in Section~\ref{con}.

\section{Related Work}
\label{rela}

In this section, we review related works from four aspects, including detection, UDA detection, alignment strategy for UDA detection and exponential moving average.

\subsection{Object Detection}

Object detection is a fundamental topic in computer vision and has been extensively studied for decades. In general, existing modern detectors can be categorized into either anchor-based or anchor-free. Anchor-based detectors usually contain a set of anchor boxes with different scales and aspect ratios, which are applied to generate object proposals for further processing (in two-stage frameworks) or final detections (in one-stage frameworks). One of the most popular anchor-based detectors is Faster R-CNN~\cite{DBLP:journals/pami/RenHG017}. It introduces a novel region proposal network (RPN) to produce object proposals based on anchors and then applies another network to further process the proposals for detection. The approaches of SSD~\cite{liu2016ssd} and YOLOv2~\cite{redmon2017yolo9000} present one-stage anchor-based detectors that strikes a good balance between accuracy and speed. Later, more excellent anchor-based detectors~\cite{dai2016r,he2017mask,lin2017feature,lin2017focal,zhang2018single,cai2018cascade} are proposed for improvements. Different from anchor-based approaches, anchor-free detectors remove the manual design of anchor boxes and directly predict the class and coordinates of objects. YOLO~\cite{redmon2016you} directly predicts the object class and position from grid cells. CornerNet~\cite{DBLP:conf/eccv/LawD18} proposes to predict the object bounding boxes as keypoint detection. The work of CenterNet~\cite{duan2019centernet} improves CornetNet by considering an extra center point.  FCOS~\cite{DBLP:conf/iccv/TianSCH19} introduces the fully convolutional networks to predict object box of each pixel in feature maps. Recently, DETR~\cite{carion2020end} applies Transformer~\cite{vaswani2017attention} to develop an anchor-free detector~ and exhibits impressive performance.

In this paper, we utilize the proposed MGA upon the popular anchor-based Faster R-CNN~\cite{DBLP:journals/pami/RenHG017} and anchor-free FCOS~\cite{DBLP:conf/iccv/TianSCH19} to verify its effectiveness. But please note that, our MGA is general and flexible and can be used in more frameworks for UDA detection.

\subsection{Unsupervised Domain Adaption (UDA) Detection}

The task of unsupervised domain adaption (UDA) detection focuses on improving the generality of object detectors learned from labeled source images on unlabeled target images. Because of its great practicability, UDA object detection has attracted extensive attention in recent years. One popular framework is to leverage adversarial learning to achieve UDA detection~\cite{DBLP:conf/cvpr/VSGOSP21,zhao2022task}. These approaches introduce a discriminator to identify which
domain the features of pixels, regions or images come from. Then, the goal is to confuse the discriminator to learn domain-invariant features for detection. In addition, many other researchers propose to apply graph methods for UDA detection~\cite{chen2021dual,xu2020cross,li2022sigma,li2022source}. These graph-based approaches propose to construct a graph based on regions or instances in an image and leverages the intra-class and inter-class relation of intra-domain and inter-domain for detection. Self-training strategy has also been explored in UDA detection~\cite{khodabandeh2019robust,li2021category,chen2022learning,yu2022sc}. The main idea of these methods is to generate high-quality or class-balanced pseudo-labels, which can be utilized to train the detection model on the unlabeled target domain. The approaches of~\cite{chen2020harmonizing,hsu2020progressive,shen2021cdtd} leverage the idea of style transfer for UDA detection. In specific, these methods reduce the discrepancy of data distribution between two domains by translating the images of source domain to target style, which improves the generalization ability of the detector. Besides the aforementioned methods, recent approaches propose to apply mean-teacher for UDA detection~\cite{deng2021unbiased,he2022cross,li2022cross}. These models adopt a teacher-student training framework to maintain the consistency of the teacher and student detector networks for boosting the generality of detection. Different from the above approaches, in this paper we tackle the UDA detection problem from a different perspective by unifying alignments of multiple granularities. 

\subsection{Alignment for UDA Detection}

Alignment of feature distribution between source and target domains has demonstrated effectiveness for UDA detection. Accordingly to the features involved, recent alignment-based approaches can be categorized into three types including pixel-, instance-, and category-level alignments.

\vspace{0.3em}
\noindent
\textbf{Pixel-level Alignment.} Pixel-level alignment focuses on aligning the pixel feature distributions of objects and background regions between two domains for UDA detection. The work of~\cite{DBLP:conf/eccv/HsuTLY20} takes into account every pixel for domain adaption and introduces a center-aware pixel-level alignment by paying more attention to foreground pixels for UDA detection. The approach of~\cite{DBLP:conf/cvpr/KimJKCK19} designs the multi-domain-invariant representation learning to encourage unbiased semantic representation through adversarial learning. The method of~\cite{hsu2020progressive} introduces an intermediate domain for progressive adaption and utilize adversarial learning for pixel-level feature alignment.

\vspace{0.3em}
\noindent
\textbf{Instance-level Alignment.} Instance-level alignment usually leverages features of regions or instances to train a domain discriminator. The approach of~\cite{cai2019exploring} explores instance relationships using mean-teacher based on Faster R-CNN~\cite{DBLP:journals/pami/RenHG017} for UDA detection. The work of~\cite{he2019multi} introduces a hierarchical framework to align both instance and domain features for detection. The work of~\cite{DBLP:conf/cvpr/Chen0SDG18} proposes to adapt Faster R-CNN on target domain images by aligning features on instance- and image-levels, exhibiting promising performance. The approach of~\cite{DBLP:conf/eccv/LiDZWLWZ20} introduces attention mechanisms for better alignment in UDA detection. Instead of using all instances or regions for alignment, the work of~\cite{zhu2019adapting} proposes to mine the discriminative ones and focuses on aligning them across two different domains for adaption detection. The work of ~\cite{Li_2023_WACV} leverages both image- and object-level adaptation to reduce the discrepancy in feature distributions, and then enforces a new domain-level metric regularization on the generated auxiliary domain by data augmentation. The method of ~\cite{Cao_2023_CVPR} further analyzes object-level features via contrastive learning based on high-quality pseudo-labels generated by other methods.

\vspace{0.3em}
\noindent
\textbf{Category-level Alignment.} Category-level alignment considers the intrinsic relation of feature distributions in source and target domains and exploits the categorical discriminability for alignment. The approaches of~\cite{DBLP:conf/iccv/DuTYFXZYZ19, DBLP:conf/cvpr/HuKSC20, DBLP:conf/eccv/PaulTSRC20, jiang2021decoupled} learn a category-specific discriminator for each category and focus on classification between two domains using pseudo labels. Despite effectiveness, it is difficult for these approaches to learn discriminative category-wise representation among multiple discriminators. The method of~\cite{DBLP:conf/eccv/WangSZD020} retains one discriminator to distinguish different categories within one domain, whereas it neglects the consistency of feature subspaces in the same category across two domains. Besides, the work of~\cite{DBLP:conf/cvpr/XuZJW20} develops a categorical regularization method that focuses on important regions and instances to reduce the domain discrepancy. The method of~\cite{DBLP:conf/cvpr/XuWNTZ20} introduces category-level domain alignment by enhancing intra-class compactness and inter-class separability and shows promising performance for domain adaption detection. 

\vspace{0.3em}
\noindent
\textbf{Our Alignment.} Despite sharing similar spirit in applying alignment for UDA object detection, our approach (\ie, Multi-Granularity Alignment (MGA)) is significantly different from others. Specifically, MGA is a unified framework that effectively encodes the dependencies across different granularities, including pixel-, instance-, and category-levels, for domain adaption detection, while other methods do not consider this important dependency relation in alignment. In addition, we specially design the omni-scale
gated fusion (OSGF) module and present a new category-level discriminator in MGA to improve the discriminative ability.

\subsection{Exponential Moving Average}

Exponential moving average (EMA) is a simple but effective strategy for updating the model parameters and commonly utilized in distillation technology~\cite{chen2020big,cai2021exponential,islam2021dynamic} and mean-teacher~\cite{xu2021end,yang2021interactive}. In these approaches, the EMA process is usually controlled by a constant weight coefficient to update the teacher network. Despite its effectiveness, this mechanism may hurt the performance of UDA detection due to the low-quality pseudo labels generated by the teacher detector. Therefore, unlike previous studies, we design an adaptive EMA (AEMA) by exploring the assessments of the teacher and student detector networks during training. AEMA can adjust the weight coefficient of EMA in an adaptive manner, leading to higher-quality pseudo labels for improvements.

\subsection{Teacher-Student Network}

Our MGA for domain adaptive object detection is based on the teacher-student framework, and there are many relevant works~\cite{Ke_2019_ICCV,ge2020mutual,Huo2023FocusOY}. For example, the method of~\cite{Ke_2019_ICCV} proposes to select stable samples by employing a stability comparison approach and conducting the stabilization constraint during the training process within a teacher-student framework, and effectively improves the performance for domain adaptive classification. The work of~\cite{ge2020mutual} aims to learn better features from the target domain with a teacher-student model for person re-identification, which is achieved via offline refined hard pseudo labels generated by the clustering-based UDA method and online refined soft pseudo labels obtained by collaboratively training two networks in an alternative manner. The work of~\cite{Huo2023FocusOY} introduces a dual teacher-student model with a bidirectional learning strategy to balance the learning and adapting abilities for better domain adaptive semantic segmentation, showing promising results.

Despite sharing a similar spirit with the above methods in applying the teacher-student model, our work is different. First, we focus on the task of domain adaptive object detection in this work and aim to reduce the discrepancy in feature distributions between two domains within our MGA method, while other works deal with different tasks. In addition, we present a simple but effective AEMA strategy to improve the quality of pseudo labels, which enhances the teacher-student model for better detection performance from a perspective different than other methods.

\begin{figure*}[!t]
\centering
\includegraphics[width=0.98\linewidth]{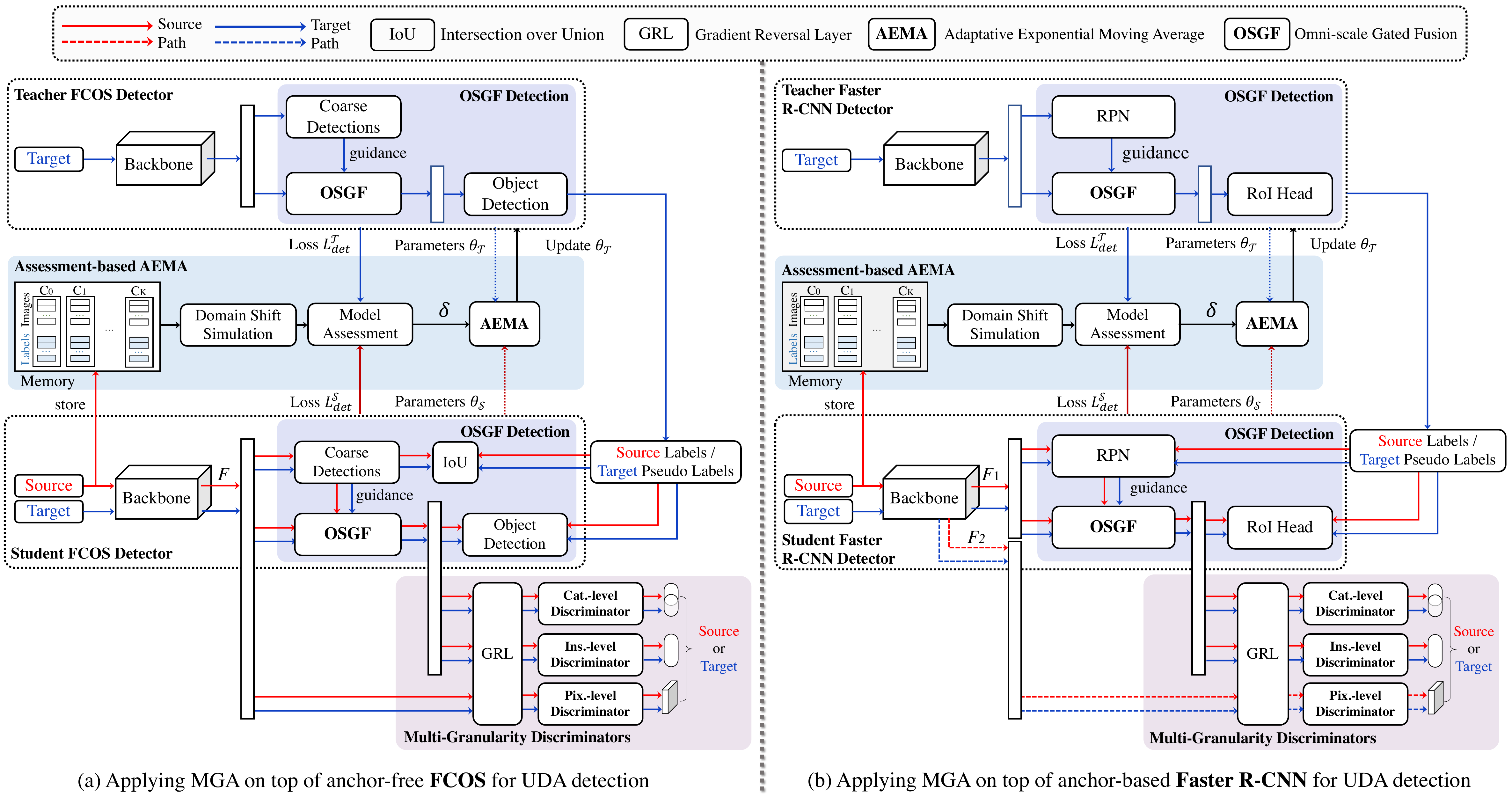}
\caption{Framework of our MGA on the top of popular anchor-free FCOS~\cite{DBLP:conf/iccv/TianSCH19} (see left image (a)) and anchor-based Faster R-CNN~\cite{DBLP:conf/cvpr/Ren0ZPC0018} (see right image (b)) for UDA detection with assessment-based AEMA. Note that for Faster R-CNN, the region proposal network (RPN) and the RoI head are used for coarse detection and final detection, respectively. $F$ in (a), $F_1$ and $F_2$ in (b) represent the features from the feature pyramid network in FCOS and backbone in Faster R-CNN.}
\label{fig:fig2}
\end{figure*}

\section{Multi-Granularity Alignment (MGA)}
\label{mga}

\subsection{Overview}

In this paper, we propose a novel unified multi-granularity alignment (MGA) framework for domain adaption detection. The overall architecture is illustrated in Figure~\ref{fig:fig2}.

As displayed in Figure~\ref{fig:fig2}, given images from the source domain $s$ and the target domain $t$, we first extract the base pixel-level feature representation from the backbone. Then, these features are merged in the omni-scale gated fusion (OSGF) module to produce discriminative representations of multi-scale instances. Based on the fused feature representations, more accurate candidate objects can be predicted by the object detection head. Meanwhile, we introduce the multi-granularity discriminators to distinguish the feature distributions between two domains from different perspectives, including pixel-level, instance-level and category-level. Moreover, in order to improve the quality of pseudo labels and mitigate the misalignment issue caused by noisy pseudo labels during training, we propose a simple but effective assessment-based adaptive EMA (AEMA) strategy to refine the pseudo labels, further enhancing the robustness of our MGA for domain adaption detection.

To achieve the proposed MGA framework,  three key parts (\i.e \ding{172}, \ding{173}, and \ding{174}) need to be designed.

It is worth noting that, our MGA is a general framework and can be easily applied in various detectors (\eg, anchor-free FCOS~\cite{DBLP:conf/iccv/TianSCH19} and anchor-based Faster R-CNN~\cite{DBLP:journals/pami/RenHG017}) with different backbones (\eg, VGG-16~\cite{DBLP:journals/corr/SimonyanZ14a} and ResNet-101~\cite{DBLP:conf/cvpr/HeZRS16}). Without loss of generality, we first apply the proposed MGA in FCOS \cite{DBLP:conf/iccv/TianSCH19} for UDA object detection as in Figure~\ref{fig:fig2} (a), and then explain how it can be used in Faster R-CNN~\cite{DBLP:journals/pami/RenHG017} as in Figure~\ref{fig:fig2} (b). For FCOS~\cite{DBLP:conf/iccv/TianSCH19}, we extract feature maps from the last three stages of the backbone and combine them into multi-level feature maps $F^k$, where $k\in\{3,4,5,6,7\}$, using FPN representation~\cite{DBLP:conf/cvpr/LinDGHHB17}. 

\subsection{Omni-Scale Gated Object Detection}
\label{sec_detection}

In most previous studies on UDA detection, the main goal is to designate discriminators at a specific level and some attentive regions. However, the use of point representation at pixel-level in anchor-free models \cite{DBLP:conf/eccv/HsuTLY20,DBLP:journals/corr/abs-2110-00249} may cause difficulties in learning robust and discriminative feature in cluttered background, while the pooling operation (\eg, RoIAlign~\cite{he2017mask}) in anchor-based models \cite{DBLP:conf/cvpr/SaitoUHS19,DBLP:conf/eccv/HeZ20} may distort features of objects with different scales and aspect ratios. 

To handle this issue, we introduce an omni-scale gated fusion (OSGF) module for object detection, which enables the adaption of the feature learning to object with various scales and aspect ratios. Specifically in OSGF, with the scale guidance from coarse detections, we can choose the most plausible convolutions with different kernels to extract compact features of instances in terms of object scales, which can significantly boost the discriminative capacity of the features. Our OSGF module is designed for general purpose and thus can be easily applied in different detectors.

\subsubsection{Scale Guidance by Coarse Detection}
\label{scale_gui}

In order to select the most plausible convolutions for feature extraction, it is necessary to obtain the scale information of the objects. To this end, we introduce a coarse detection step to provide the scale guidance. In specific, followed by the multi-level feature maps $F^{k}$ ($k\in\{3,4,5,6,7\}$ denotes the level index) from the backbone (see Figure~\ref{fig:fig2} (a)), we can predict the candidate object boxes $\tilde{b}^{k}$ through a series of convolutional layers. Drawing inspiration from~\cite{DBLP:conf/cvpr/RezatofighiTGS019}, we utilize the cross-entropy Intersection over Union (IoU) loss \cite{DBLP:conf/mm/YuJWCH16} to regress the bounding boxes of objects in foreground pixels as follows,
\begin{equation} \label{eq_candidate}
\mathcal{L}_\text{gui} = -\sum_k\sum_{(i,j)}\ln(\text{IoU}(\tilde{b}^k_{i,j}, b^k_{i,j})),
\end{equation}
where $\text{IoU}(\cdot,\cdot)$ represents the function to calculate the IoU score between predicted box $\tilde{b}^{k}$ and ground-truth box $b^{k}$. For each pixel $(i, j)$ in the feature map, the corresponding box $b^{k}_{i,j}$ can be defined as a $4$-dimensional vector as follows,

\begin{equation}
b^{k}_{i,j}=(x_{t_{i,j}}, x_{b_{i,j}}, x_{l_{i,j}}, x_{r_{i,j}})
\end{equation}
where $x_{t_{i,j}}$, $x_{b_{i,j}}$, $x_{l_{i,j}}$, and $x_{r_{i,j}}$ respectively represent the distances between current location and the top, bottom, left and right bounds of ground-truth box. Therefore, the normalized object scale (\ie, width $w^k$ and height $h^k$) at each level can be computed as follows,
\begin{equation}\label{eq_scale}
\left\{
\begin{aligned}
&w^k_{i,j} = (\tilde{x}_{r_{i,j}}+\tilde{x}_{l_{i,j}})/\text{stride}^k,\\
&h^k_{i,j} = (\tilde{x}_{b_{i,j}}+\tilde{x}_{t_{i,j}})/\text{stride}^k,\\
\end{aligned}
\right.
\end{equation}
where $\text{stride}^k$ denotes how many steps we move in each round of convolution operation\footnote{We have $\{(k,\textit{stride})|(3, 8), (4, 16), (5, 32), (6, 64), (7, 128)\}$.}. As in FCOS~\cite{DBLP:conf/iccv/TianSCH19}, the feature maps at each level are utilized to individually detect the objects of different scales in the range $\{[-1, 64]$, $[64, 128]$, $[128, 256]$, $[256, 512]\}$, $[512, +\infty]\}$. Therefore, the majority of object scales is less than $8$, \ie, $w^k\leq8, h^k\leq8$. For notation simplicity, we omit the superscript $k$ and write $F$ for $F^k$ and $\tilde{b}$ for $\tilde{b}^k$ in the following sections.

\begin{figure*}[!t]
\centering
\includegraphics[width=\linewidth]{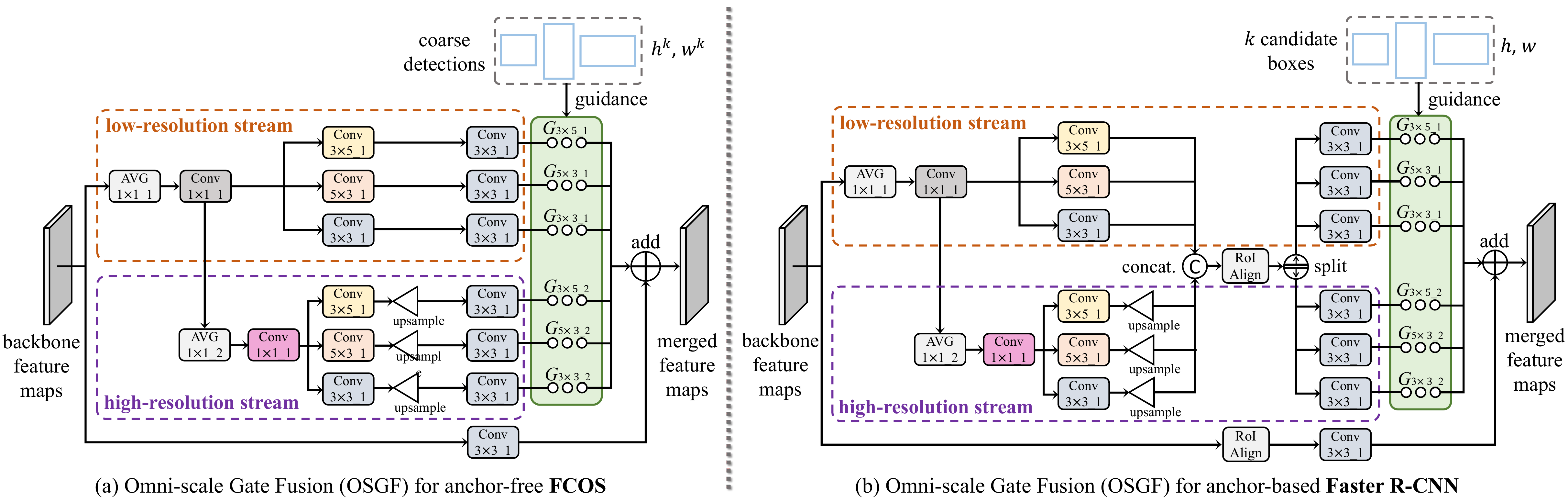}
\caption{Illustration of the proposed omni-scale gated fusion (OSGF) module for anchor-free FCOS~\cite{DBLP:conf/iccv/TianSCH19} (see left image (a)) and anchor-based Faster R-CNN~\cite{DBLP:conf/cvpr/Ren0ZPC0018} (see right image (b)). The parameters of
the modules with the same color are shared.}
\label{fig:fig3}
\end{figure*}

\subsubsection{Omni-scale Gated Fusion (OSGF)} 

With the scale guidance as in Section~\ref{scale_gui}, we present an omni-scale gated fusion module (OSGF), which is composed of both \textit{low-resolution} and \textit{high-resolution} feature streams, to adapt to objects of various scales and aspect ratios. Specifically, as illustrated in Figure \ref{fig:fig3}, the low-resolution stream consists of three parallel convolutional layers with different kernels $\omega\in\{3\times3,3\times5,5\times3\}$, which is applied for feature extraction of relatively small objects ($w^k\leq5, h^k\leq5$). Meanwhile, in the high-resolution feature stream, we use another set of parallel convolutional layers with kernels $\omega$ to handle large objects ($w^k>5, h^k>5$). The different from the low-resolution feature branch is that, we utilize an extra upsampling operation after each convolutional layer in the high-resolution stream to upscale the feature maps. It is worth noticing that, the structure of OSGF in this paper is different from that in the conference publication~\cite{zhou2022multi}. In specific, the major modifications in this paper include: (i) removal of the $3\times 3$ convolutional layer before the two streams, (ii) incorporation of simpler averaging pooling and $1\times 1$ convolutional layers in the low-resolution stream, (iii) replacement of the $3\times 3$ convolutional layers with shared averaging pooling and $1\times 1$ convolutional layers, and (iv) change of the $1\times 1$ convolutional layer to $3\times 3$ convolutional layer in the residual connection. By doing so, the overall number of parameters is significantly reduced because of less convolutional layers used. In addition, we observe that the detection performance has been improved with our overall structure of the OSGF module, as evidenced by our experiments in Tab.~\ref{tab_modified_gate} described later. It is worth noting that, the designed shared convolutional weights in OSGF in this work are to reduce its number of parameters and simplify its overall architecture, which makes the proposed MGA more concise and less heavy. Likewise, as shown in Tab.~\ref{tab_modified_gate}, this shared design has effectively reduced the number of parameters of OSGF. More details will be illustrated in later experiments.

After the two branches of low- and high-resolution features, we introduce a gate mask $G$ to weight each convolutional layer based on the predicted coarse boxes $\tilde{b}$ as follows,

\begin{equation}\label{eq_mask}
G_{\omega} = \frac{\exp(\tau(o_{\omega}-\hat{o})/(\hat{o}+\epsilon))}{\sum_{\omega}\exp(\tau(o_{\omega}-\hat{o})/(\hat{o}+\epsilon))},
\end{equation}
where $\tau$ represents the temperature factor, $o_{\omega}=\text{IoU}(\tilde{b}, \omega)$ denotes the overlap between the predicted box and the convolution kernel $\omega$, and $\hat{o}$ is the maximal overlap among them. 
Finally, we merge the pixel-level features to exploit the scale-wise representation of instances as follows,
\begin{equation}\label{eq_feature}
M = \sum_{\omega}F_{\omega}\odot G_{\omega}+F_{3\times 3},
\end{equation}
where $\odot$ denotes the element-wise product, and $F_\omega$ denotes feature maps after the convolutional layer with kernel $\omega$.

\subsubsection{Object Detection} 
After obtaining the merged feature maps $M$ from the OSGF module, we can predict the categories and bounding boxes of objects. In FCOS~\cite{DBLP:conf/iccv/TianSCH19}, the object detection heads contain three branches for classification, centerness and regression, respectively. The classification and centerness branches are optimized by the focal loss \cite{DBLP:conf/iccv/LinGGHD17} $\mathcal{L}_\text{cls}$ and cross-entropy loss \cite{DBLP:conf/iccv/TianSCH19} $\mathcal{L}_\text{ctr}$, respectively. The regression branch is optimized by the IoU loss \cite{DBLP:conf/mm/YuJWCH16} $\mathcal{L}_\text{reg}$. Thus, the final loss function $\mathcal{L}_\text{det}$ for the object detection  is defined as
\begin{equation} 
\mathcal{L}_\text{det} = \mathcal{L}_\text{cls} + \mathcal{L}_\text{ctr} + \mathcal{L}_\text{reg}.
\label{eqn_det_loss}
\end{equation}
Please refer to~\cite{DBLP:conf/iccv/TianSCH19} for more details regarding the loss functions. It is worthy to notice that, in the UDA detection, we implement two detectors, including a teacher detector and a student detector (see Figure~\ref{fig:fig2} (a)). These two detectors share the same architecture but independent parameters. We denote the loss functions for the teacher and the student detectors as $\mathcal{L}_\text{det}^{\mathcal{T}}$ and $\mathcal{L}_\text{det}^{\mathcal{S}}$, respectively.

\subsection{Multi-Granularity Discriminators}

As discussed earlier, we propose the multi-granularity discriminators to distinguish  whether the sample belongs to the source domain or the target domain from various perspectives, consisting of pixels, instances and categories. The discrepancy between two domains is reduced using Gradient Reversal Layer (GRL)~\cite{DBLP:conf/icml/GaninL15} that transfers reverse gradient when optimizing the object detection network. The discriminator contains four stacked convolution-groupnorm-relu layers and an extra $3\times3$ convolutional layer. Below we will elaborate on our multi-granularity discriminators.

\subsubsection{Pixel- and Instance-level Discriminators} 

The pixel- and instance-level discriminators are leveraged to respectively perform pixel-level and instance-level alignments of feature maps between two domains. As demonstrated in Figure \ref{fig:fig2} (a), given the input multi-level features $F$ and the merged feature $M$, the pixel-level and instance-level discriminators $D^\text{pix}$ and $D^\text{ins}$ are learned through the loss functions $\mathcal{L}_\text{pix}$ and $\mathcal{L}_\text{ins}$. Similar to previous work~\cite{DBLP:conf/eccv/HsuTLY20}, we adopt the same loss function. Then, $\mathcal{L}_\text{pix}$ and $\mathcal{L}_\text{ins}$ are defined as follows
\begin{equation} \label{eq:dis_pix}
\begin{split}
\mathcal{L}_\text{pix} = -\sum_{(i,j)} & y^\text{pix}_{i,j}\log D^\text{pix}(F^s(i,j)) \\
       &+ (1-y^\text{pix}_{i,j})\log(1-D^\text{pix}(F^t(i,j))),
\end{split}
\end{equation}
\begin{equation} \label{eq:dis_ins}
\begin{split}
\mathcal{L}_\text{ins} = -\sum_{(i,j)} & y^\text{ins}_{i,j}\log D^\text{ins}(M^s(i,j)) \\
       &+ (1-y^\text{ins}_{i,j})\log(1-D^\text{ins}(M^t(i,j))),
\end{split}
\end{equation}
where $F(i,j)$ is the feature at pixel $(i,j)$ in $F$, and $M(i,j)$ the feature at instance $(i,j)$ in $M$. We have the domain label $y^\text{pix}_{i,j}=1$ if pixel at $(i,j)$ in $F$ is from source domain and $0$ otherwise. Likewise, $y^\text{ins}_{i,j}=1$ if the instance at $(i,j)$ in $M$ belongs to source domain and $0$ otherwise.

\begin{figure}[t]
\centering
\includegraphics[width=0.9\linewidth]{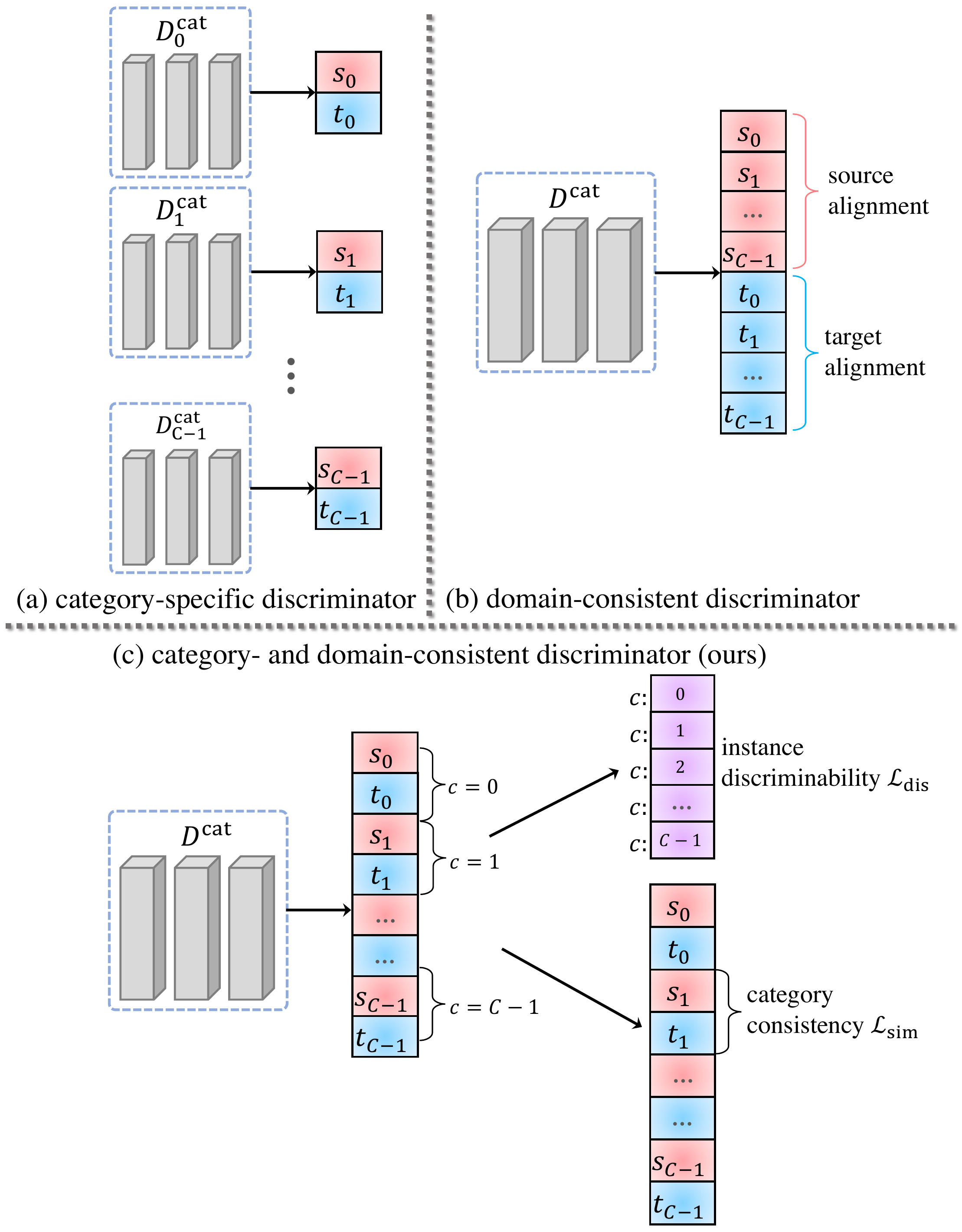}
\caption{Illustration of different category-level discriminators $D$. $s_c$ and $t_c$ are the $c$-th category ($c=0,1,\cdots,C-1$) in source and target domains respectively. (a) Category-specific discriminators for each category \cite{DBLP:conf/iccv/DuTYFXZYZ19, DBLP:conf/cvpr/HuKSC20, DBLP:conf/eccv/PaulTSRC20}. (b) Domain-consistent discriminator to distinguish different categories within one domain~\cite{DBLP:conf/eccv/WangSZD020}. (c) Our category- and domain-consistent discriminator to consider both instance discriminability in different categories and category consistency between two domains.}
\label{fig:discriminator}
\end{figure}

\subsubsection{Category-level Discriminator}
\label{cat_dis}

In order to keep the semantic consistency between different domain distributions, a category-level discriminator is applied. Previous methods design either category-specific discriminators for each category (\eg,~\cite{DBLP:conf/iccv/DuTYFXZYZ19, DBLP:conf/cvpr/HuKSC20, DBLP:conf/eccv/PaulTSRC20}, see Figure~\ref{fig:discriminator} (a)) or a domain-consistent discriminator to distinguish  categories within one domain (\eg,~\cite{DBLP:conf/eccv/WangSZD020}, see Figure~\ref{fig:discriminator} (b)). By contrast, our approach considers jointly instance discriminability in different categories and category consistency between two domains and introduces a novel category- and domain-consistent discriminator (see Figure~\ref{fig:discriminator} (c)). Specifically, in our discriminator, we predict the category and domain labels of pixel $(i,j)$ in each image based on feature map $\hat{M}\in \mathbb{R}^{H\times W\times 2C}$, where $\hat{M}\in \mathbb{R}^{H\times W\times 2C}$ is the output by feeding $M$ to the category-level discriminator, $H$ and $W$ are the height and width respectively, and $2C$ represents the total number of categories for source and target domains. 

Since there is no ground-truth to supervise the category-level discriminator, we assign pseudo labels to important samples with high confidence from object detection (see Sec. \ref{sec_detection}). In practice, given a batch of input images, we can output the category probability map $P$ using the object detection heads, and obtain the set $\mathcal{S}$ of pseudo labels by utilizeing the probability threshold $\tau_\text{prob}$ and non-maximum suppression (NMS) threshold $\tau_\text{nms}$. Then, the instances in different categories are classified by Eq. \eqref{eq_dis_cls}, while the same category in two domains is aligned by Eq. \eqref{eq_sim_cls}, as follows:
\begin{itemize}
\item In order to keep \textbf{\textit{instance discriminability in different categories}}, we separate the category distribution by using the following loss function,
\begin{equation} \label{eq_dis_cls}
\mathcal{L}_\text{dis} = -\frac{1}{|\mathcal{S}|}\sum_{(i,j)\in \mathcal{S}}\sum_{c=0}^{C-1}\hat{y}^\text{dis}_{i,j,c}\log(p^\text{dis}_{i,j,c}).
\end{equation}
By normalizing confidence over the domain channel, $p_{i,j,c}^\text{dis}$ represents the probability of the $c$-th category of the pixel $(i,j)$, \ie,
\begin{equation} \label{eq_cls_prob}
\small
p^\text{dis}_{i,j,c} = \frac{\exp{(\hat{M}_{i,j,2c}+\hat{M}_{i,j,2c+1})}}{\sum_{c=0}^{C-1}\exp{(\hat{M}_{i,j,2c}+\hat{M}_{i,j,2c+1})}},
\end{equation}
where $\hat{M}_{i,j,2c}$ and $\hat{M}_{i,j,2c+1}$ represent the confidence of the $c$-th category in source and target domains, respectively (see again Figure \ref{fig:discriminator}(c)). $\hat{y}^\text{dis}\in\mathbb{R}^{H\times W\times C}$ is the pseudo category label. We have $\hat{y}^\text{dis}_{i,j,c}=1$ if the instance at $(i,j)$ in $\hat{M}$ is an important one of the $c$-th category and $\hat{y}^\text{dis}_{i,j,c}=0$ otherwise.

\item \textbf{\textit{Category consistency between two domains.}} After classifying instances of different categories, we need to further identify which domain the instance belongs to. With GRL~\cite{DBLP:conf/icml/GaninL15}, we write the loss function as follows,
\begin{equation} \label{eq_sim_cls}
    \mathcal{L}_\text{sim} = -\frac{1}{|\mathcal{S}|}\sum_{(i,j)\in\mathcal{S}}\sum_{m=0}^{2C-1}\hat{y}^\text{sim}_{i,j,m}\log(p^\text{sim}_{i,j,m}),
\end{equation}
where $y^\text{sim}\in\mathbb{R}^{H\times W\times 2C}$ is the pseudo domain label. Similarly, $\hat{y}^\text{sim}_{i,j,m}=1$ if the instance at $(i,j)$ in $\hat{M}$ is an important one of the $\lfloor\frac{m}{2}\rfloor$-th category in specific domain and $\hat{y}^\text{sim}_{i,j,m}=0$ otherwise. The domain probability $p^\text{sim}$ is obtained as follows,

\begin{equation} \label{eq_sim_prob}
    p^\text{sim}_{i,j,m} = \left\{
    \begin{array}{ll}
        \frac{\exp(\hat{M}_{i,j,m})} {\exp(\hat{M}_{i,j,m-1}) + \exp(\hat{M}_{i,j,m})},  & {\rm if~} m {\rm ~is~odd}  \\
        \frac{\exp(\hat{M}_{i,j,m})} {\exp(\hat{M}_{i,j,m}) + \exp(\hat{M}_{i,j,m+1})}.  & {\rm if~} m {\rm ~is~even}
    \end{array}
    \right.
\end{equation}
\end{itemize}

With the above analysis, we define the final loss function $\mathcal{L}_\text{cat}$ for the category-level discriminator $D^\text{cat}$ as follows,
\begin{equation} \label{eq_cls}
\mathcal{L}_\text{cat} = \lambda_\text{dis}\mathcal{L}_\text{dis} + \lambda_\text{sim}\mathcal{L}_\text{sim},
\end{equation}
where $\mathcal{L}_\text{dis}$ and $\mathcal{L}_\text{sim}$ are loss functions for instance discriminability and category consistency as in Eq. \eqref{eq_dis_cls} and Eq. \eqref{eq_sim_cls}, and $\lambda_\text{dis}$ and $\lambda_\text{sim}$ are the balancing factors.

\subsection{Adaptive Exponential Moving Average (AEMA)} 

As mentioned in Section~\ref{cat_dis}, the pseudo labels, which are generated by the teacher detector (see Figure~\ref{fig:fig2} (a)), are required for supervising the learning of the category-level discriminator. During the training procedure, the teacher detector is usually updated using exponential moving average (EMA) as follows,
\begin{equation}\label{ema}
\theta_{\mathcal{T}}^{\eta} = (1 - \gamma) \cdot \theta_{\mathcal{T}}^{\eta-1} + \gamma \cdot \theta_{\mathcal{S}}^{\eta-1}
\end{equation}
where $\theta_{\mathcal{T}}^{\eta}$ represents the weights of the teacher detector at iteration $\eta$, $\theta_{\mathcal{S}}^{\eta-1}$ denotes the weights of the student detector at iteration $\eta-1$, and $\gamma$ is a constant coefficient.

\begin{figure}[!t]
\centering
\includegraphics[width=\linewidth]{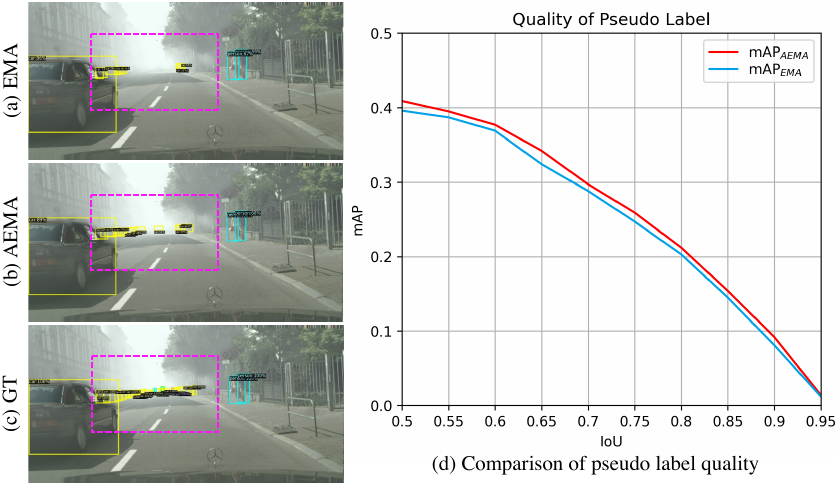}
\caption{Comparison of the pseudo label quality between EMA and AEMA. Image (a) displays the pseudo label generated by EMA, image (b) the pseudo label by our AEMA, and image (c) the GT pseudo label. In image (d), we show mAP scores of the generated pseudo labels of different strategies, and we can see that AEMA produces better pseudo labels.}
\label{fig:fig5}
\end{figure}

Despite simplicity, the EMA approach may lead to some low-quality pseudo labels (see Figure~\ref{fig:fig5} (a)), because it does not consider the feedback from the two detectors, degrading the final detection performance. To address this problem for improving quality of pseudo labels, we propose an adaptive EMA (AEMA). Specifically, unlike EMA, AEMA considers the intermediate assessments of both teacher and student detectors during update by evaluating their performance on the source domain. Using the assessments as a guidance, an update factor $\delta$ (as described later) is learned to adjust the coefficient $\alpha$ in Eq. (\ref{ema}). More concretely, as in Figure~\ref{fig:fig2}, we maintain a memory bank, which is used for generating the assessments. The memory is dynamically updated by storing the images and labels of source domain into it. In order to accurately assess the generalization ability of the detector, we introduce a domain shift simulation (DSS, see Figure.~\ref{fig:fig2} (a)) module, and apply it on the memory bank to generate discrepancy of data distribution on the source domain. Specifically, given the sampling data of images $x^{m}$ and labels $y^{m}$ of all categories from the memory bank, we randomly adjust the mean $x_{u}$ and variance $\sigma^{2}$ of $x^{m}$ to generate the variant data distribution from the source domain, as follows
\begin{equation} \label{eq_sim_prob_1}
\overline{x}^{m} = \widetilde{\sigma} \frac{x^{m} - x_{u}}{\sqrt{\sigma^{2} + \epsilon}} + \widetilde{x}_{u},
\end{equation}
Where the mean $\widetilde{x}_{u}$ and the standard deviation $\widetilde{\sigma}$ for the new variant data distribution are obtained by using the uniform distribution under $x_{u}$ and $\sigma$, respectively, as follows, 
\begin{equation} \label{eq_sim_prob_2}
\widetilde{x}_{u} = U(a_{u},b_{u})x_{u},
\end{equation}
\begin{equation} \label{eq_sim_prob_3}
\widetilde{\sigma} = U(a_{\sigma},b_{\sigma})\sigma,
\end{equation}
Here, $U(a,b)$ represents the uniform distribution between $a$ and $b$ and is predefined. Afterwards, the detection losses $\mathcal{L}_\text{det}^{\mathcal{T}}$ and $\mathcal{L}_\text{det}^{\mathcal{S}}$, obtained by evaluating the teacher and student detectors on the above input date, are employed as the assessment results to derive the update factor $\delta$, as follows,
\begin{equation} 
\delta = 
\begin{cases}
    e^{\tau_{1}\cdot(0.5-\rho)}, \rho < 0.5\\
    e^{\tau_{2}\cdot(0.5-\rho)}, \rho \ge 0.5
\end{cases} \;\;
\rho = \frac{L_{det}^{\mathcal{S}}}{L_{det}^{\mathcal{S}} + L_{det}^{\mathcal{T}}}
\end{equation}
where $\tau_{1}$ and $\tau_{2}$ are two constant values.

Finally, the weights of the teacher detection model can be updated by our AEMA with $\delta$ as follows,
\begin{equation}\label{ema1}
\theta_{\mathcal{T}}^{\eta} = (1 - \gamma\cdot\delta) \cdot \theta_{\mathcal{T}}^{\eta-1} + \gamma \cdot\delta \cdot \theta_{\mathcal{S}}^{\eta-1}
\end{equation}

By using AEMA, we can take into account assessments of two detectors to guide the update of teacher detection, leading to better pseudo labels, as in Figure~\ref{fig:fig5} (b). Furthermore, we show the statistic comparison of the pseudo labels obtained by teacher detector with EMA and our proposed AEMA in term of accuracy in Figure~\ref{fig:fig5} (d). As demonstrated in Figure~\ref{fig:fig5} (d), we can see that, the quality of the pseudo labels is clearly improved. We will further analyze the effectiveness of our AEMA in later experimental section. 

\subsection{Overall Loss Function and Optimization}

As discussed above, the omni-scale gated object detection network is supervised by $\mathcal{L}_\text{gui}$ and $\mathcal{L}_\text{det}$. Meanwhile, the multi-granularity discriminators are optimized in different granularities, including pixel-level $\mathcal{L}_\text{pix}$, instance-level $\mathcal{L}_\text{ins}$ and category-level $\mathcal{L}_\text{cat}$. In summary, the overall loss function is defined as
\begin{equation} \label{eq_overall}
\mathcal{L} = (\underbrace{\mathcal{L}_\text{gui} + \mathcal{L}_\text{det}}_{\text{object detection}}) \ + \alpha \hspace{-2mm} \underbrace{(\mathcal{L}_\text{pix} + \mathcal{L}_\text{ins} + \mathcal{L}_\text{cat})}_{\text{multi-granularity discriminators}}
\end{equation}
where $\alpha$ is the balancing factor between object detection and multi-granularity discriminators.

The training process of our proposed method is divided into two stages. In stage 1 (S1), we train teacher detector by using SGD optimizer and random sampling with Eq. (\ref{eqn_det_loss}) on source domain. Next, in stage 2 (S2), the student detector is optimized by using SDG optimizer and Eq. (\ref{eq_overall}) on source with labels and target domains with pseudo labels, and the teacher detector is updated by AEMA.

\begin{table*}[t]
\footnotesize
\setlength{\tabcolsep}{6pt}
\center
\caption{Results of our approach and comparison to state-of-the-arts on weather adaptation from Cityscapes to FoggyCityscapes. The best two results are shown in \textbf{bold} and \emph{italic} fonts, respectively. Note that, MGA-DA~\cite{zhou2022multi} is the method from our conference version.}
\begin{tabular}{rccccccccccc}
\hline
Method &Detector & Backbone & person & rider & car & truck & bus & train & mbike & bicycle & mAP  \\
\hline
Baseline &Faster-RCNN &VGG-16  & 17.8 & 23.6 & 27.1 & 11.9 & 23.8 & 9.1 & 14.4 & 22.8 & 18.8 \\
DAF \cite{DBLP:conf/cvpr/Chen0SDG18} &Faster-RCNN &VGG-16 &25.0 &31.0 &40.5 &22.1 &35.3 &20.2 &20.0 &27.1 &27.6 \\
SC-DA \cite{DBLP:conf/cvpr/ZhuPYSL19} &Faster-RCNN &VGG-16 & 33.5 & 38.0 & 48.5 & 26.5 & 39.0 & 23.3 & 28.0 & 33.6 & 33.8 \\
MAF \cite{DBLP:conf/iccv/HeZ19} &Faster-RCNN &VGG-16 &28.2 &39.5 &43.9 &23.8 &39.9 &33.3 &29.2 &33.9 &34.0 \\
SW-DA \cite{DBLP:conf/cvpr/SaitoUHS19} &Faster-RCNN &VGG-16 & 29.9 & 42.3 & 43.5 & 24.5 & 36.2 & 32.6 & 30.0 & 35.3 & 34.3 \\
DAM \cite{DBLP:conf/cvpr/KimJKCK19} &Faster-RCNN &VGG-16  & 30.8 & 40.5 & 44.3 & 27.2 & 38.4 & 34.5 & 28.4 & 32.2 & 34.6 \\
MOTR \cite{DBLP:conf/cvpr/CaiPNTDY19} &Faster-RCNN &ResNet-50 &30.6 &41.4 &44.0 &21.9 &38.6 &40.6 &28.3 &35.6 &35.1\\
CST \cite{DBLP:conf/eccv/ZhaoLXL20} &Faster-RCNN &VGG-16 &32.7 &44.4 &50.1 &21.7 &45.6 &25.4 &30.1 &36.8 &35.9\\
PD \cite{wu2021instance} &Faster-RCNN &VGG-16 &33.1 &43.4 &49.6 &22.0 &45.8 &32.0 &29.6 &37.1 &36.6\\
CDN \cite{DBLP:conf/eccv/SuWZTCQW20} &Faster-RCNN &VGG-16 &35.8 &45.7 &50.9 &30.1 &42.5 &29.8 &30.8 &36.5 &36.6\\
SFOD-Masoic-Defoggy \cite{DBLP:journals/corr/abs-2012-05400} &Faster-RCNN &VGG-16 &34.1 &44.4 &51.9 &30.4 &41.8 &25.7 &30.3 &37.2 &37.0\\
ATF \cite{DBLP:conf/eccv/HeZ20} &Faster-RCNN &VGG-16  & 34.6 & 46.5 & 49.2 & 23.5 & 43.1 & 29.2 & 33.2 & 39.0 & 37.3  \\
SW-Faster-ICR-CCR \cite{DBLP:conf/cvpr/XuZJW20} &Faster-RCNN &VGG-16 &32.9 &43.8 &49.2 &27.2 &45.1 &36.4 &30.3 &34.6 &37.4 \\
SCL \cite{DBLP:journals/corr/abs-1911-02559} &Faster-RCNN &VGG-16 &31.6 &44.0 &44.8 &30.4 &41.8 &40.7 &33.6 &36.2 &37.9 \\
CFFA \cite{DBLP:conf/cvpr/Zheng0LW20} &Faster-RCNN &VGG-16  & 43.2 & 37.4 & 52.1 & 34.7 & 34.0 &46.9 & 29.9 & 30.8 & 38.6 \\
GPA \cite{DBLP:conf/cvpr/XuWNTZ20} &Faster-RCNN &ResNet-50 &32.9 &46.7 &54.1 &24.7 &45.7 &41.1 &32.4 &38.7 &39.5\\
SAPNet \cite{DBLP:conf/eccv/LiDZWLWZ20} &Faster-RCNN &VGG-16 &40.8 &46.7 &59.8 &24.3 &46.8 &37.5 &30.4 &40.7 &40.9 \\
DSS \cite{DBLP:conf/cvpr/Wang_2021} & Faster-RCNN & ResNet-50 &42.9& 51.2 &53.6 &33.6 &49.2 &18.9 &36.2 &41.8& 40.9\\
D-adapt \cite{jiang2021decoupled} & Faster-RCNN & VGG-16 &44.9 &\emph{54.2} &\emph{61.7} &25.6 &36.3 &24.7 &37.3 &\emph{46.1} &41.3 \\
UMT \cite{DBLP:conf/cvpr/Deng0CD21} &Faster-RCNN &VGG-16 &\textbf{56.5} &37.3 &48.6 &30.4 &33.0 &46.7 &\emph{46.8} &34.1 &41.7 \\
MeGA-CDA \cite{DBLP:conf/cvpr/VSGOSP21} &Faster-RCNN &VGG-16 &37.7 &49.0 &52.4 &25.4 &49.2 &\emph{46.9} &34.5 &39.0 &41.8 \\
CDG \cite{li2021category} &Faster-RCNN &VGG-16 &38.0 &47.4 &53.1 &34.2 &47.5 &41.1 &38.3 &38.9 &42.3 \\
TIA \cite{zhao2022task} & Faster-RCNN & VGG-16 & \emph{52.1} & 38.1 & 49.7 & \textbf{37.7} & 34.8 & 46.3 & \textbf{48.6} & 31.1 & 42.3 \\
SDA \cite{DBLP:journals/corr/Qianyu21} & Faster-RCNN & VGG-16 & 38.3 & 47.2 & 58.8 & \emph{34.9} & \textbf{57.7} & \textbf{48.3} & 35.7 & 42.0 & \emph{45.2} \\
TDD \cite{he2022cross} & Faster-RCNN & VGG-16 & 39.6 & 47.5 & 55.7 & 33.8 & 47.6 & 42.1 & 37.0 & 41.4 & 43.1 \\
MGA-DA \cite{zhou2022multi} & Faster-RCNN & VGG-16 & 43.9 & 49.6 & 60.6 & 29.6 & 50.7 & 39.0 & 38.3 & 42.8 & 44.3 \\
\hline
\rowcolor{gray!10} Baseline (ours) &Faster-RCNN &VGG-16 & 39.4 & 46.8 & 48.2 & 23.2 & 36.0 & 16.8 & 35.2 &  43.2 & 36.1 \\
\rowcolor{gray!20} MGA (ours) &Faster-RCNN &VGG-16 &47.0 & \textbf{54.6} & \textbf{64.8} & 28.5 & \emph{52.1} & 41.5 & 40.9 & \textbf{49.5}  & \textbf{47.4}\\
\hline
oracle &Faster-RCNN &VGG-16  & 48.2 & 53.3 & 68.5 & 31.7 & 55.3 & 33.1 & 41.9 & 49.3 & 47.7 \\

\hline\hline
SST-AL \cite{DBLP:journals/corr/abs-2110-00249} &FCOS &- &45.1 & 47.4 &59.4 & 24.5 &50.0 &25.7 &26.0 &38.7 & 39.6  \\
CFA \cite{DBLP:conf/eccv/HsuTLY20} &FCOS &VGG-16  &41.9 & 38.7 & 56.7 & 22.6 & 41.5 & 26.8 & 24.6 & 35.5 & 36.0  \\
SCAN \cite{DBLP:conf/aaai/Li_2022} &FCOS & VGG-16 & 41.7 & 43.9 & 57.3 & 28.7 & 48.6 & 48.7 & 31.0 & 37.3 & 42.1 \\
SIGMA \cite{li2022sigma} & FCOS & VGG-16 & 46.9 & 48.4 & 63.7 & 27.1 & 50.7 & 35.9 & 34.7 & 41.4 & 43.5 \\
MGA-DA \cite{zhou2022multi} & FCOS  & VGG-16  & 45.7 &47.5 &60.6 &31.0 & 52.9 & 44.5 &29.0 & 38.0  & 43.6\\
CFA \cite{DBLP:conf/eccv/HsuTLY20} &FCOS &ResNet-101 &41.5 & 43.6 &57.1 & 29.4 &44.9 &39.7 &\emph{29.0} & 36.1 & 40.2 \\
MGA-DA \cite{zhou2022multi} & FCOS  & ResNet-101  & 43.1 & 47.3 &61.5 & 30.2 &53.2 &\emph{50.3} & 27.9 & 36.9 &43.8\\

\hline
\rowcolor{gray!10} Baseline (ours) & FCOS  & VGG-16  & 31.4 & 31.0 & 42.7 & 14.2 & 26.9 & 2.3 &  17.6 & 31.6 & 24.7 \\
\rowcolor{gray!10} Baseline (ours) & FCOS  & ResNet-101  & 35.2 & 37.2 & 43.5 & 17.3 & 31.8 & 7.2 & 27.0 & 34.5  & 29.2 \\
\rowcolor{gray!20} MGA (ours) & FCOS  & VGG-16  & \textbf{47.9} & \textbf{50.1} & \textbf{64.9} & \emph{34.8} & \textbf{58.0} & 45.6 & \textbf{38.3} &  \emph{43.7} & \textbf{47.9} \\
\rowcolor{gray!20} MGA (ours) & FCOS  & ResNet-101 & \emph{47.2} & \emph{48.1} & \emph{63.7} & \textbf{37.5} & \emph{54.6} & \textbf{50.8} & 28.8 & \textbf{44.2} & \emph{46.9}\\
\hline
oracle &FCOS &VGG-16  & 51.7 & 48.9 & 69.5 & 39.1 & 52.8 & 56.0 & 31.8 & 40.0 & 48.7  \\
oracle &FCOS &ResNet-101 & 46.3 & 46.0 & 66.8 & 38.8 & 57.3 & 52.2 & 36.4 & 36.9 & 47.6 \\
\hline
\end{tabular}
\label{tab_city_foggy}
\end{table*}

\section{Experiments}
\label{exp}

{\bf Extension of our framework.} Our MGA framework is designed for general purpose and applicable to both one- and two-stage detection models. To verify this, in addition to the representative one-stage FCOS~\cite{DBLP:conf/iccv/HeZ19}, we further extend our MGA to the popular two-stage Faster-RCNN~\cite{DBLP:journals/pami/RenHG017} that consists of  Region Proposal Network (RPN) and RCNN with classification and regression branches. As shown in Figure~\ref{fig:fig2} (b), we employ RPN as our coarse detection module, whose loss function is replaced by the original RPN loss, \ie, $L_\text{gui} = L_\text{rpn}$; and we use RCNN as object detection module with the loss defined as $L_\text{det} = L_\text{cls} + L_\text{reg}$. For omni-scale gate fusion, we first obtain the top $K$ proposals by using RPN based on the backbone feature layer with stride 16. Then, we further extract the pixel-level features with low-resolution and high-resolution streams, and generate instance features of $7 \times 7$ under the pixel-level feature map and original input feature maps by using the ROIAlign operation. Finally, the instance features are merged according to the RPN outputs and Eq. (\ref{eq_feature}) after using a convolution of $3 \times 3$, as shown in Figure~\ref{fig:fig3} (b).

\vspace{0.3em}
\noindent
{\bf Implementation.}
In this work, we implement our method based on different detectors (\ie, Faster-RCNN and FCOS) and backbones (\ie, VGG-16 and ResNet-101) using PyTorch \cite{paszke2019pytorch} to show generality of our approach. Both VGG-16 and ResNet-101 are pre-trained on ImageNet \cite{krizhevsky2012imagenet}. We utilize a unified optimization framework by using training process of two stages and warm-up followed the previous works \cite{DBLP:conf/eccv/HsuTLY20} and \cite{li2022sigma} for different detectors. Similar to~\cite{DBLP:conf/eccv/LiDZWLWZ20}, we apply the Adam optimizer with an initial learning rate of 3e-4, a momentum of 0.9 and weight decay of 1e-4 in Faster-RCNN framework. For FCOS framework, we use SGD optimizer with an initial learning rate of 5e-3, a momentum of 0.9 and weight decay of 1e-4, being consistent with CFA \cite{DBLP:conf/eccv/HsuTLY20}. $\gamma$ is $0.1$ in Eq. (\ref{ema}). The parameters $a_{u}$ and $b_{u}$ are set respectively to $0.4$ and $0.5$ in Eq. (\ref{eq_sim_prob_2}), and the $a_{\delta}$ and $b_{\delta}$ respectively to $0.8$ and $0.9$ in Eq. (\ref{eq_sim_prob_3}). The thresholds   $\tau_\text{prob}$ and $\tau_\text{nms}$ for obtaining $\mathcal{S}$ are empirically set to $0.42$ and $0.5$. All our experiments are conducted on the machine with an Intel(R) Xeon(R) CPU and 4 Tesla V100 GPUs. Our code will be made publicly available at  \url{https://github.com/tiankongzhang/MGA}.

\subsection{Datasets}
To verify the proposed method, we conduct extensive experiments on different adaption settings, as described below.

\begin{table*}[t]
\footnotesize
\setlength{\tabcolsep}{5.5pt}
\center
\caption{Results of our approach and comparison to state-of-the-arts on real-to-artistic adaptation from PASCAL VOC to Clipart. The best two results are
shown in \textbf{bold} and \emph{italic} fonts, respectively. Note, there are no oracle results for Clipart because all images in Clipart are used for evaluation.}
\begin{tabular}{rccccccccccccc}
\hline
Method & Detector & Backbone & acro & bicycle & bird & boat & bottle & bus & car & cat & chair & cow &   \\
\hline
Baseline &Faster-RCNN & ResNet-101  & 35.6 & 52.5 & 24.3 & 23.0 & 20.0 & 43.9 & 32.8 & 10.7 & 30.6 & 11.7 & \\

SW-DA \cite{DBLP:conf/cvpr/SaitoUHS19} & Faster-RCNN & ResNet-101 & 26.2 & 48.5 & 32.6 & 33.7 & 38.5  & 54.3 & 37.1 & 18.6 & 34.8 & 58.3 \\
SCL \cite{DBLP:journals/corr/abs-1911-02559} & Faster-RCNN & ResNet-101 & \emph{44.7} & 50.0 & 33.6 & 27.4 & 42.2 & 55.6 & 38.3 & 19.2 & 37.9 & \emph{69.0} \\
ATF \cite{DBLP:conf/eccv/HeZ20} & Faster-RCNN & ResNet-101 & 41.9 & 67.0 & 27.4 & 36.4 & 41.0 & 48.5 & 42.0 & 13.1 & 39.2 & \textbf{75.1} \\
PD \cite{wu2021instance} & Faster-RCNN & ResNet-101 & 41.5 & 52.7 & 34.5 & 28.1 & 43.7 & 58.5 & 41.8 & 15.3 & 40.1 & 54.4 \\
SAPNet \cite{DBLP:conf/eccv/LiDZWLWZ20} & Faster-RCNN & ResNet-101 & 27.4 & \emph{70.8} & 32.0 & 27.9 & 42.4 & 63.5 & 47.5 & 14.3 & \textbf{48.2} & 46.1 \\
UMT \cite{deng2021unbiased} & Faster-RCNN & ResNet-101 & 39.1 & 59.1 & 32.4 & 35.0 & 45.1 & 61.9 & 48.4 & 7.5  & \emph{46.0} & 67.6\\
SFOD-ODS \cite{li2022source} & Faster-RCNN & ResNet-101 & 43.1 & 61.4 & \emph{40.1} & 36.8 & \emph{48.2} & 45.8 & 48.3 & 20.4  & 44.8 & 53.3 \\
D-adapt \cite{jiang2021decoupled} & Faster-RCNN & ResNet-101 & \textbf{56.4} & 63.2 & \textbf{42.3} & \textbf{40.9} & 45.3 & \emph{77.0} & \emph{48.7} & \emph{25.4} & 44.3 &  58.4 \\
MGA-DA \cite{zhou2022multi} & Faster-RCNN & ResNet-101 & 35.5 & 64.6 & 27.8 & 34.5 & 41.6 & 66.4 & \textbf{49.8} & \textbf{26.8} & 43.6 & 56.7 \\
\hline
\rowcolor{gray!10} Baseline (ours) &Faster-RCNN & ResNet-101 & 30.5 & 35.3 & 24.8 & 23.5 & 34.8 & 65.7 & 32.6  & 9.0  & 35.1 & 26.4 &  \\
\rowcolor{gray!20} MGA (ours) &Faster-RCNN & ResNet-101 & 38.7 & \textbf{77.2} & 39.0 & \emph{35.4} & \textbf{53.8} & \textbf{78.1} &  47.5 & 17.5  & 38.2 & 49.9 &  \\
\hline \hline
 & &  & table & dog & horse & bike & person & plant & sheep & sofa & train & tv & mAP \\
 Baseline & Faster-RCNN & ResNet-101  & 13.8 & 6.0 & 36.8 & 45.9 & 48.7 & 41.9 & 16.5 & 7.3 & 22.9 & 32.0 & 27.8 \\
 SW-DA \cite{DBLP:conf/cvpr/SaitoUHS19} & Faster-RCNN & ResNet-101 & 17.0 & 12.5 & 33.8 & 65.5 & 61.6 & 52.0 & 9.3 & 24.9 & 54.1 & 49.1 & 38.1 \\
 SCL \cite{DBLP:journals/corr/abs-1911-02559} & Faster-RCNN & ResNet-101 & 30.1 & 26.3 & 34.4 & 67.3 & 61.0 & 47.9 & 21.4 & 26.3 & 50.1 & 47.3 & 41.5 \\
 ATF \cite{DBLP:conf/eccv/HeZ20} & Faster-RCNN & ResNet-101 & \textbf{33.4} & 7.9 & 41.2 & 56.2 & 61.4 & 50.6 & 42.0 & 25.0 & 53.1 & 39.1 & 42.1 \\
 PD \cite{wu2021instance} & Faster-RCNN & ResNet-101 & 26.7 & \emph{28.5} & 37.7 & 75.4 & 63.7 & 48.7 & 16.5 & 30.8 & 54.5 & 48.7 & 42.1 \\
 SAPNet \cite{DBLP:conf/eccv/LiDZWLWZ20} & Faster-RCNN & ResNet-101 & 31.8 & 17.9 & 43.8 & 68.0 & 68.1 & 49.0 & 18.7 & 20.4 & 55.8 & 51.3 & 42.2 \\
 UMT \cite{deng2021unbiased} & Faster-RCNN & ResNet-101 & 21.4 & \textbf{29.5} & \textbf{48.2} & 75.9 & 70.5 & \emph{56.7}  & 25.9 & 28.9 & 39.4 & 43.6 & 44.1 \\
 SFOD-ODS \cite{li2022source} & Faster-RCNN & ResNet-101 & \emph{32.5} & 26.1 & 40.6 & \textbf{86.3} & 68.5 &  48.9 & 25.4 & \emph{33.2} & 44.0 & 56.5 & 45.2 \\
 D-adapt \cite{jiang2021decoupled} & Faster-RCNN & ResNet-101 &  31.4 & 24.5 & \emph{47.1} & 75.3 & 69.3 & 43.5 & \textbf{27.9}  & \textbf{34.1} & \textbf{60.7} & \textbf{64.0} & \textbf{49.0} \\
 MGA-DA \cite{zhou2022multi} & Faster-RCNN & ResNet-101 & 24.3 & 20.9 & 43.2 &  \emph{84.3} & \emph{74.2} & 41.1 & 17.4 & 27.6 & \emph{56.5} & 57.6 & 44.8 \\
 \hline
 \rowcolor{gray!10} Baseline (ours) & Faster-RCNN & ResNet-101  & 24.2 & 12.2 & 31.2 & 55.5 & 40.4 & 52.2 & 5.7 & 18.4 & 45.0 & 38.4  & 32.0 \\
 \rowcolor{gray!20} MGA (ours) & Faster-RCNN & ResNet-101  & 20.0 & 18.0 & 44.2 & 83.5 & \textbf{74.6} & \textbf{57.7} & \emph{26.7} & 26.0 & 55.4 & \emph{58.3}  & \emph{47.0} \\
\hline
\end{tabular}
\label{tab_city_clipart}
\end{table*}

\begin{table*}[t]
\footnotesize
\setlength{\tabcolsep}{11pt}
\center
\caption{Results of our approach and comparison to state-of-the-arts on real-to-artistic adaptation from PASCAL VOC to Watercolor. The best two results are
shown in \textbf{bold} and \emph{italic} fonts, respectively.}
\begin{tabular}{rccccccccc}
\hline
Method & Detector & Backbone & bike & bird & car & cat & dog & person & mAP\\
\hline
Baseline &Faster-RCNN & ResNet-101  & 68.8 & 46.8 & 37.2 & 32.7 & 21.3 & 60.7 & 44.6  \\
SW-DA \cite{DBLP:conf/cvpr/SaitoUHS19} & Faster-RCNN & ResNet-101 & 82.3 & 55.9 & 46.5 & 32.7 & 35.5 & 66.7 & 53.3 \\
SCL \cite{DBLP:journals/corr/abs-1911-02559} & Faster-RCNN & ResNet-101 & 82.2 & 55.1 & 51.8 & 39.6 & 38.4 & 64.0 & 55.2 \\
ATF \cite{DBLP:conf/eccv/HeZ20} & Faster-RCNN & ResNet-101 & 78.8 & \textbf{59.9} & 47.9 & 41.0 & 34.8 & 66.9 & 54.9 \\
PD \cite{wu2021instance} & Faster-RCNN & ResNet-101 & \textbf{95.8} & 54.3 & 48.3 & \emph{42.4} & 35.1 & 65.8 & 56.9 \\
SAPNet \cite{DBLP:conf/eccv/LiDZWLWZ20} & Faster-RCNN & ResNet-101 & 81.1 & 51.1 & 53.6 & 34.3 & 39.8 & 71.3 & 55.2 \\
UMT \cite{deng2021unbiased} & Faster-RCNN & ResNet-101 & 88.2 & 55.3 & 51.7 & 39.8 & 43.6 &  69.9 & 58.1 \\
SFOD-ODS \cite{li2022source} & Faster-RCNN & ResNet-101 & \emph{95.2} & 53.1 & 46.9 & 37.2 & \textbf{47.6} & 69.3 & 58.2 \\
AT \cite{li2022cross} & Faster-RCNN & ResNet-101 & 93.6 & 56.1 & \emph{58.9} & 37.3 & 39.6 &  \emph{73.8} & \emph{59.9} \\
MGA-DA \cite{zhou2022multi} & Faster-RCNN & ResNet-101 & 87.6 & 49.9 & 56.9 & 37.4 & 44.6 & 72.5 & 58.1 \\
\hline
\rowcolor{gray!10} Baseline (ours) &Faster-RCNN & ResNet-101 & 76.0 & 46.7 & 52.0 & 27.7 & 33.3 & 54.9 & 48.4 \\
\rowcolor{gray!20} MGA (ours) &Faster-RCNN & ResNet-101 & 85.3 & \emph{59.7} & \textbf{59.7} & \textbf{43.3} & \emph{46.6} & \textbf{77.7} & \textbf{62.1}  \\
\hline
oracle &Faster-RCNN & ResNet-101 & 67.1 & 53.4 & 43.9 & 46.3 & 50.5 & 79.8 & 56.8 \\
\hline
\end{tabular}
\label{tab_city_watercolor}
\end{table*}

\vspace{0.3em}
\noindent
\textbf{Weather adaptation.} For weather adaptation, we explore generalization of the detector on Cityscapes~\cite{DBLP:conf/cvpr/CordtsORREBFRS16} and  FoggyCityscapes~\cite{DBLP:journals/ijcv/SakaridisDG18}. Cityscapes~\cite{DBLP:conf/cvpr/CordtsORREBFRS16} is a popular street scene dataset with normal weather, which comprises 2,975 training images and 500 validation images. FoggyCityscapes~\cite{DBLP:journals/ijcv/SakaridisDG18} is synthesized on Cityscapes with different levels of fog (\ie, 0.005, 0.01 and 0.02). For fair comparison, we choose the level of $0.02$ for experiment as in other methods in Table~\ref{tab_city_foggy}. In weather adaptation, we use Cityscapes~\cite{DBLP:conf/cvpr/CordtsORREBFRS16} as the source domain and FoggyCityscapes~\cite{DBLP:journals/ijcv/SakaridisDG18} as the target domain.

\vspace{0.3em}
\noindent
\textbf{Cross-Camera adaptation.} In Cross-Camera adaptation, we evaluate our algorithm on KITTI \cite{DBLP:/conf/cvpr/are12} and Cityscapes. KITTI \cite{DBLP:/conf/cvpr/are12} is a popular traffic scene dataset containing 7,481 training images. In this adaption experiment, KITTI is the source domain and Cityscapes is the target domain. Following~\cite{DBLP:journals/corr/abs-1911-02559,li2022sigma}, we only report results on the category of car. 

\vspace{0.3em}
\noindent
\textbf{Synthetic-to-Real adaptation.} For Synthetic-to-Real adaptation, we utilize SIM10k \cite{DBLP:/conf/icra/driving17} and Cityscapes for experiments. SIM10k \cite{DBLP:/conf/icra/driving17} is a synthetic scene dataset from the game video Grand Theft Auto V (GTA5). It contains 10k training images, and we conduct comparisons on the car class, similar to~\cite{DBLP:journals/corr/abs-1911-02559}. In this adaptation experiment, we utilize the SIM10k as the source domain and Cityscapes as the target domain.

\vspace{0.3em}
\noindent
\textbf{Real-to-Artistic adaptation.} In Real-to-Artistic adaptation, we verify our method on PASCAL VOC \cite{DBLP:journals/ijcv/EveringhamGWWZ10}, Clipart \cite{DBLP:conf/cvpr/InoueFYA18} and Watercolor \cite{DBLP:conf/cvpr/InoueFYA18} datasets. PASCAL VOC \cite{DBLP:journals/ijcv/EveringhamGWWZ10} is a real-scene dataset including two sub-datasets (\ie, PASCAL VOC 2007 and PASCAL VOC 2012). PASCAL VOC 2007 consists of 2,501 images for training and 2,510 images for validation, and PASCAL VOC 2012 contains 5,717 images for training and 5,823 mages for validation. Clipart \cite{DBLP:conf/cvpr/InoueFYA18} is a carton dataset with 1k images and has the same categories as PASCAL VOC. Watercolor \cite{DBLP:conf/cvpr/InoueFYA18} is a watercolor style dataset containing 1,000 training images and 1,000 testing images, and it shares 6 classes with PASCAL VOC. In this setting, we use PASCAL VOC as the source domain and Clipart or Watercolor as the target domain.

\subsection{State-of-the-Art Comparison}

\begin{figure*}[!t]
\centering
\includegraphics[width=0.93\linewidth]{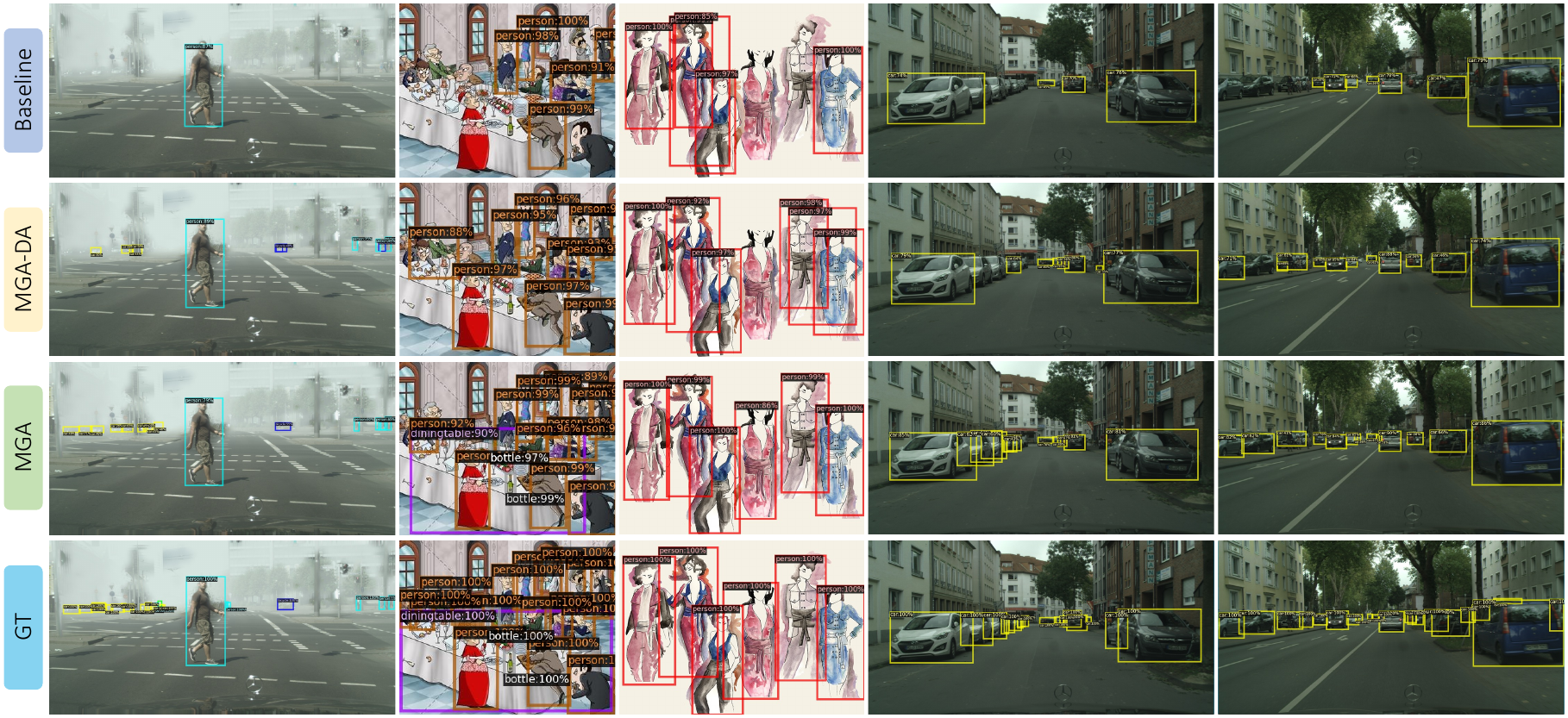}
\caption{Qualitative results and comparison (from left column to right column: weather adaptation from Cityscapes to FoggyCityscapes, real-to-artistic adaptation from PASCAL VOC to Clipart and Watercolor, cross-camera adaption from Kitti to Cityscapes, and synthetic-to-real adaption from SIM10k to Cityscapes). We can see that MGA achieves superior results than MGA-DA in our conference version and the baseline method.}
\label{fig:vis}
\end{figure*}

\begin{table}[!t]
\small
\setlength{\tabcolsep}{3pt}
\centering
\caption{Results of our approach and comparison to state-of-the-arts on cross-camera/synthetic-to-real adaptation detection results from Kitti/SIM10k to Cityscapes. The best two results are
displayed in \textbf{bold} and \emph{italic} fonts, respectively.
}
\begin{tabular}{rccc}
\hline
Method &Detector & Backbone & AP$_\text{car}$ \\
\hline
Baseline &Faster-RCNN &VGG-16  & 30.2/30.1 \\
DAF \cite{DBLP:conf/cvpr/Chen0SDG18} &Faster-RCNN &VGG-16  & 38.5/39.0 \\
MAF \cite{DBLP:conf/iccv/HeZ19} &Faster-RCNN&VGG-16  & 41.0/41.1 \\
ATF \cite{DBLP:conf/eccv/HeZ20} &Faster-RCNN &VGG-16  & 42.1/42.8 \\
UMT \cite{DBLP:conf/cvpr/Deng0CD21} &Faster-RCNN &VGG-16 & -/43.1 \\
SFOD-Mosaic \cite{DBLP:journals/corr/abs-2012-05400} &Faster-RCNN   &VGG-16   & 44.6/43.1 \\
CST \cite{DBLP:conf/eccv/ZhaoLXL20} &Faster-RCNN   &VGG-16   & 43.6/44.5 \\
MeGA-CDA \cite{DBLP:conf/cvpr/VSGOSP21} &Faster-RCNN &VGG-16 & 43.0/44.8 \\
SAPNet \cite{DBLP:conf/eccv/LiDZWLWZ20}&Faster-RCNN   &VGG-16   & 43.4/44.9 \\
CDN \cite{DBLP:conf/eccv/SuWZTCQW20} &Faster-RCNN   &VGG-16   & 44.9/49.3 \\
TIA \cite{zhao2022task}  &Faster-RCNN & VGG-16 & 44.0/\: -- \: \\
DSS \cite{DBLP:conf/cvpr/Wang_2021}  &Faster-RCNN & ResNet-50 &42.7/44.5  \\
GPA \cite{DBLP:conf/cvpr/XuWNTZ20}  &Faster-RCNN & ResNet-50 &\emph{47.9}/47.6  \\
SSD \cite{DBLP:conf/iccv/Rezaeianaran_2021}  &Faster-RCNN & ResNet-50 &47.6/49.3  \\
TDD \cite{he2022cross}  &Faster-RCNN & VGG-16 & 47.4/\emph{53.4}  \\
MGA-DA \cite{zhou2022multi}  &Faster-RCNN & VGG-16 & 45.2/49.8  \\
\hline
\rowcolor{gray!10} Baseline (ours) &Faster-RCNN &VGG-16  & 43.7/44.1 \\
\rowcolor{gray!20} MGA (ours) &Faster-RCNN &VGG-16  & \textbf{54.3}/\textbf{55.5} \\
\hline
oracle &Faster-RCNN &VGG-16  & 68.1  \\
\hline\hline
SST-AL \cite{DBLP:journals/corr/abs-2110-00249} &FCOS &- & 45.6/51.8 \\
CFA \cite{DBLP:conf/aaai/Li_2022} &FCOS   &VGG-16   & 43.2/49.0 \\
SCAN \cite{DBLP:conf/aaai/Li_2022} &FCOS   &VGG-16   &  45.8/52.6 \\
SIGMA \cite{li2022sigma}  &FCOS & VGG-16 & 45.8/53.7  \\
CFA \cite{DBLP:conf/eccv/HsuTLY20} &FCOS  &ResNet-101  & 45.0/51.2 \\
MGA-DA \cite{zhou2022multi} &FCOS   &VGG-16   & \emph{48.5}/\emph{54.6} \\
MGA-DA \cite{zhou2022multi} &FCOS  &ResNet-101  & 46.5/54.1 \\
\hline
\rowcolor{gray!10} Baseline (ours) & FCOS  & VGG-16  &  43.1/43.0 \\
\rowcolor{gray!10} Baseline (ours) & FCOS  & ResNet-101  &  41.3/43.7  \\
\rowcolor{gray!20} MGA (ours) & FCOS  & VGG-16  & \textbf{49.9}/\textbf{55.8} \\
\rowcolor{gray!20} MGA (ours) & FCOS  & ResNet-101   &  47.6/55.4  \\
\hline
oracle &FCOS  &VGG-16  &73.4   \\
oracle &FCOS  &ResNet-101  &71.8   \\
\hline
\end{tabular}
\label{tab_sim10k_city}
\end{table}

In this section, we demonstrate our results  and comparison with state-of-the-art methods using different base detectors (\ie, Faster-RCNN \cite{DBLP:journals/pami/RenHG017} and FCOS \cite{DBLP:conf/iccv/TianSCH19}) and backbones (\ie, VGG-16 \cite{DBLP:journals/corr/SimonyanZ14a} and ReseNet-101 \cite{DBLP:conf/cvpr/HeZRS16}) on different adaptation scenarios. In all comparison table, ``Baseline (ours)'' means that the baseline detector is equipped with our OSGF and trained using data augmentation as in our method but without adaption strategy, and ``oracle'' indicates that the baseline detector is trained and tested on the target domain without any adaptation strategy.

\vspace{0.3em}
\noindent
\textbf{Weather adaptation.} In Table \ref{tab_city_foggy}, we report the results from Cityscapes to FoggyCityscapes. As displayed in Table~\ref{tab_city_foggy}, for Faster-RCNN detector, our method achieves the best mAP of $47.4\%$ and outperforms the second best SDA \cite{DBLP:journals/corr/Qianyu21} with $45.2\%$ mAP by $2.4\%$. Compared to the baseline (ours) of $36.1\%$, MGA obtains $11.3\%$ performance gains. For FCOS detector, we also obtain the best mAP scores of $47.9\%$ with VGG-16 and $46.9\%$ with ResNet-101. Compared with the approaches of SIGMA \cite{li2022sigma} with VGG-16 and CFA \cite{DBLP:conf/eccv/HsuTLY20} with ResNet-101, our MGA respectively shows the $4.4\%$ and $6.7\%$ gains. In addition, our method observes obvious improvements over the baseline (ours) with $23.2\%$ using VGG-16 and $17.7\%$ gains using ResNet-101, which verifies the effectiveness of our method. Furthermore, compared with MGA-DA~\cite{zhou2022multi} with $44.3\%$ mAP score for Fatser-RCNN and $43.6\%$ and $43.8 \%$ mAP scores for FCOS on VGG-16 and ResNet-101 backbone in our conference version, our MGA shows $3.1\%$, $4.3\%$ and $3.1\%$ gains, showing the effectiveness of our new contributions.

\vspace{0.3em}
\noindent
\textbf{Real-to-Artistic adaptation.} Table \ref{tab_city_clipart} and \ref{tab_city_watercolor} show the results and comparison in real-to-artistic adaptation. As in Table \ref{tab_city_clipart}, from PASCAL VOC to Clipart, our method achieves the second mAP score of $47.0\%$, and D-adapt \cite{jiang2021decoupled} obtains the best performance of $49.0\%$. Compared to our baseline, we achieve $15.0\%$ performance gain. In Table \ref{tab_city_watercolor}, our MGA performs the best with $62.1\%$ mAP score and surpasses the second best AT \cite{li2022cross} with $59.9\%$ by $2.2\%$. Besides, our method outperforms the oracle, indicating that our MGA makes full use of the information between source and target domains for robust UDA detection.

\vspace{0.3em}
\noindent
\textbf{Cross-Camera adaptation.} Table \ref{tab_sim10k_city} displays results and comparison from Kitti to Cityscapes. As shown in Table  \ref{tab_sim10k_city}, on Faster-RCNN detector, our method shows the best result of $54.3\%$ AP$_\text{car}$ with VGG-16. In contrast to the baseline (ours), we obtain $10.6\%$ gain. Using FCOS detector, our MGA outperforms SIGMA \cite{li2022sigma} by $4.1\%$ with VGG-16 and CFA \cite{DBLP:conf/eccv/HsuTLY20} by $2.6\%$ with ResNet-101. Compared to the baseline (ours), MGA obtains $6.8\%$ and $6.3\%$ performance gains with VGG-16 and ResNet-101, respectively, showing its advantages.

\vspace{0.3em}
\noindent
\textbf{Synthetic-to-Real adaptation.} Table~\ref{tab_sim10k_city} shows the result from Sim10k to Cityscapes. On Faster-RCNN, our MGA achieves the best result of $55.5\%$ AP$_\text{car}$. In contrast to baseline (ours), it obtains a $11.4\%$ gain. On FCOS, our method shows the best AP$_\text{car}$ of $55.8\%$ with VGG-16 and $55.4\%$ with ResNet-101. In comparison with SIGMA \cite{li2022sigma} with VGG-16 and CFA \cite{DBLP:conf/eccv/HsuTLY20} with ResNet-101, we obtains $2.1\%$ and $4.2\%$ gains with VGG-16 and ResNet-101, respectively. Compared to baselines (ours), it demonstrates gains of $12.8\%$ and $11.7\%$ with VGG-16 and ResNet-101, respectively.

Besides quantitative results, we demonstrate qualitative results our method and comparison to other approaches in Figure~\ref{fig:vis}. From Figure~\ref{fig:vis}, we can observe that MGA achieves superior detection results than MGA-DA in our conference version and the baseline method.

\begin{figure}[!t]
	\centering
\includegraphics[width=\linewidth]{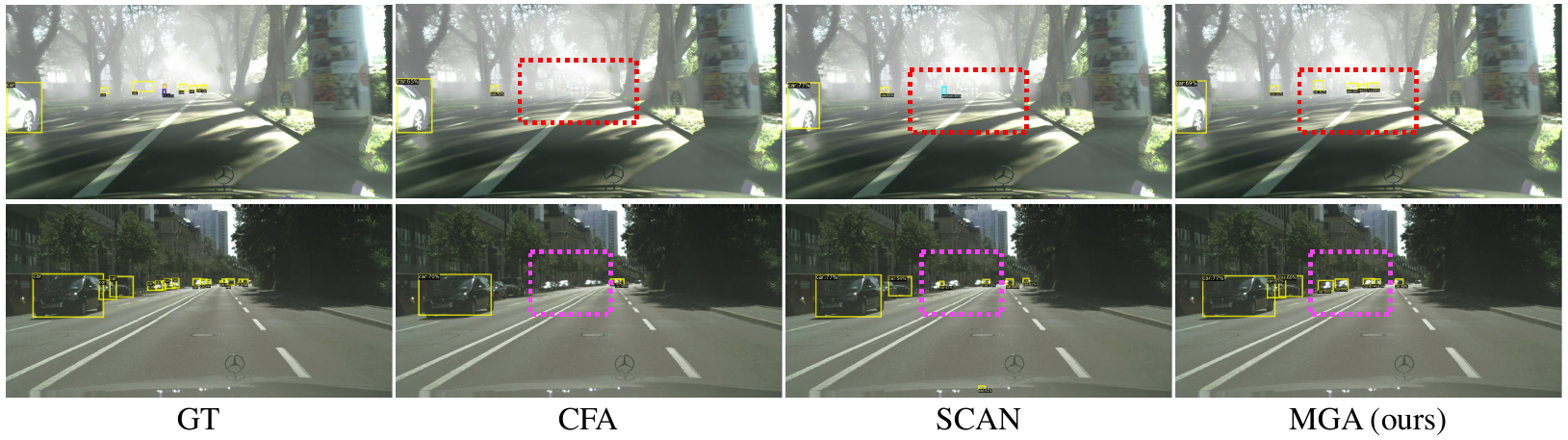}
	\caption{Qualitative comparisons of MGA with other state-of-the-art CFA and SCAN on challenging cases in different settings. The first row shows the comparison for domain adaption Cityscapes$\rightarrow$FoggyCityscapes where the fog is thick, and the second row displays comparison on Sim10k$\rightarrow$Cityscapes where the object styles largely differ. We can clearly see that other methods fail in these challenging cases while our approach effectively detects the objects, evidencing its efficacy.}
	\label{fig:rev-fig5}
\end{figure}

To further demonstrate the superiority of our approach, we further provide qualitative comparisons with state-of-the-arts in some corner cases where the scenarios are challenging. Specifically, we show the results of our MGA with state-of-the-art CFA~\cite{DBLP:conf/eccv/HsuTLY20} and SCAN~\cite{DBLP:conf/aaai/Li_2022} on the settings of Cityscapes$\rightarrow$FoggyCityscapes and Sim10k$\rightarrow$Cityscapes based on FOCS. The reason for such choice is that these methods directly provide accessible pretrained models of their approaches for evaluation and thus are convenient for comparison. Fig.~\ref{fig:rev-fig5} shows the comparison. In specific, the first two rows show the results for domain adaption Cityscapes$\rightarrow$FoggyCityscapes where the fog in the images are thick. Although these cases are challenging to CFA and SCAN (see column 2 and 3 of the first two rows), our MGA can detect most of the objects (see column 4 of the first two rows). In the last two rows of Fig.~\ref{fig:rev-fig5}, we compare our method with CFA ans SCAN on Sim10k$\rightarrow$Cityscapes where object styles largely differ. Compared to CFA and SCAN (see column 2 and 3 of the last two rows), the proposed MGA is able to effectively detect the objects from these challenging situations (see column 4), which evidences the superiority of our method in dealing with corner cases.

\subsection{Ablation Study}

To further analyze our approach, we conduct ablation experiments on different components. The results are reported under the weather adaptation.

\vspace{0.3em}
\noindent
\textbf{Effectiveness of different components.} In order to further validate the effectiveness of different components including omni-scale gated fusion (OSGF), multi-granularity discriminators (MGD) and adaptive exponential moving average (AEMA) in MGA, we demonstrate the results by gradually adding them to the baseline, which is FCOS with VGG16. Table~\ref{tab:all_modules} shows the results. From Tab~\ref{tab:all_modules}, we can see that, OSGF improves the mAP of the baseline from 22.0\% to 24.7\% with a gain of 2.7\% (\ding{202} v.s. \ding{203}). In addition, the performance of AP$^S$, AP$^M$, and AP$^L$ in detecting small, medium, and large objects are improved from 2.1\% to 3.2\%, 24.6\% to 29.9\%, and 54.3\% to 54.8\%, respectively. This shows that, even without other components, OSGF can enhance the detection in the target domain. Further, when removing OSGF from MGA (\ding{206} v.s. \ding{208}), the overall mAP score is decreased from 47.9\% to 45.6\%. Besides, the detection performance for small, medium, and large objects are also degenerated, which shows the importance of OSGF in our MGA. With MGD for distribution alignment across different granularities, the mAP score is further significantly increased from 24.7\% to 45.3\% with an obvious gain of 20.6\% (\ding{203} v.s. \ding{204}), which is the major improvement for the proposed method. Besides mAP, the scores of AP$^S$, AP$^M$, and AP$^L$ are also largely improved with MGD. Moreover, when adopting AEMA (\ding{204} v.s. \ding{208}), the final detection performance is improved from 45.3\% to 47.9\%, which evidences the improvement obtained with better pseudo labels by AEMA. In addition, we show the results of our method using MGD and EMA, without (see \ding{205}) and with (see \ding{207}) OSGF. By comparing EMA and AEMA (\ding{205} v.s. \ding{206} without OSGF and \ding{207} v.s. \ding{208} with OSGF), the mAP scores are improved from 44.4\% to 45.6\% and from 46.5\% to 47.9\%, respectively, which again confirms the effectiveness and advantage of the proposed AEMA in improving detection.

\begin{figure}[!t]
	\centering
	\includegraphics[width=\linewidth]{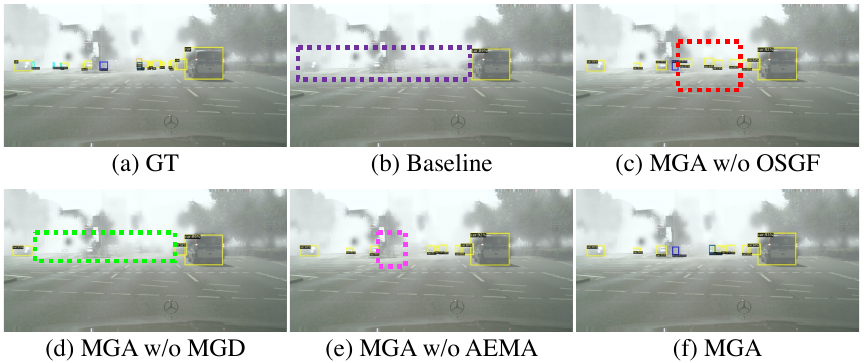}
	\caption{Qualitative results for different variants of our method.}
	\label{fig:rev-fig2}
\end{figure}

Besides the above quantitative analysis, we further clarify the improvements from each component by showing qualitative detection results, as in Fig.~\ref{fig:rev-fig2}. From Fig.~\ref{fig:rev-fig2} (c) and (f), we can see that OSGF is able to help MGA detect the objects with small scales. From Fig.~\ref{fig:rev-fig2} (d) and (f), we observe that, MGD leads to significant improvement of our MGA in detecting object from the target domain. From Fig.~\ref{fig:rev-fig2} (e) and (f), the detection quality is further enhanced owing to better pseudo labels by AEMA. All these analysis show the improvements from different components of MGA.


\vspace{0.3em}
\noindent
\textbf{Comparison of category-level discriminators.} As displayed in the Table \ref{tab:aba_dis_cls},  we compare our category-level discriminator $D^\text{cat}$ and other related class-level discriminators, including $D^\text{cen}$~\cite{DBLP:conf/aaai/Li_2022}, $D^\text{grp}$~\cite{DBLP:conf/cvpr/HuKSC20} and $D^\text{cls}$~\cite{DBLP:conf/eccv/WangSZD020} using FCOS with VGG16. The ``baseline'' indicates that the detector is learned under all strategies but the class-level discriminator is removed from MGA module. $D^\text{cen}$ focuses on reducing the differences on the center-aware distributions of source and target domains, which is composed of features of the central positions of objects. $D^\text{grp}$ aligns the feature distributions by building a sole domain discriminator for each category. $D^\text{cls}$ simultaneously takes domain and class information and expands the binary domain labels by inserting the binary class labels. 

\begin{table}[!t]
\footnotesize
\centering
\tabcolsep=0.18cm
\caption{Effectiveness of different components. Following COCO~\cite{lin2014microsoft}, AP$^\text{S}$, AP$^\text{M}$ and AP$^\text{L}$ denote the mAP scores such that object area is in the range $[0,32^2], (32^2,96^2]$, and $(96^2, +\infty)$, respectively.}
\begin{tabular}{ccccccccc}
        \hline 
        & OSGF & MGD & EMA & AEMA  & mAP & AP$^S$ & AP$^M$ & AP$^L$\\ 
        \hline 
        \ding{202}  &             &            &        &             & 22.0 &2.1 & 24.6 & 54.3 \\ 
        \ding{203}  & \checkmark  &            &        &             & 24.7 & 3.2 & 29.9 & 54.8 \\
        \ding{204}  & \checkmark  & \checkmark &        &             & 45.3 & 15.4 & 44.8 & 75.6 \\
        \red{\ding{205}}  &             & \HF{\checkmark} & \HF{\checkmark} &         & \HF{44.4} & \HF{11.1} & \HF{44.9} & \HF{75.2} \\
        \ding{206}  &             & \checkmark  &       & \checkmark  & 45.6 & 12.2 & 46.2 & 76.1 \\
        \red{\ding{207}}  & \HF{\checkmark}  & \HF{\checkmark}  &   \HF{\checkmark}    &   & \HF{46.5} & \HF{15.8} & \HF{46.3} & \HF{74.7} \\
        \ding{208}  & \checkmark  & \checkmark  &       &  \checkmark & 47.9 & 16.2 & 48.1 & 77.8 \\ 
        \hline  
        \end{tabular}
  \label{tab:all_modules}%
\end{table}

\begin{table}[!t]\footnotesize
\centering
\caption{Comparison between different category-level discriminators.}
\begin{tabular}{cccccc}
    \hline 
     discriminator & baseline & $D^\text{cen}$ & $D^\text{grp}$ & $D^\text{cls}$ & $D^\text{cat}$ (ours) \\ 
    \hline 
       mAP   &  44.7   &  46.2  & 46.0 &  45.2  & 47.9 \\ 
    \hline 
    \end{tabular}
  \label{tab:aba_dis_cls}%
\end{table}

\begin{table}[!t]\footnotesize
    \centering
    \caption{Effectiveness of modules in MGD.}
    \begin{tabular}{rc}
        \hline 
                   & mAP \\ 
        \hline 
         Baseline (no MGD)  & 24.7  \\ 
         MGA w/ $D^{\text{pix}}$ in MGD     & 40.6  \\
         MGA w/ $D^{\text{pix}}$ and $D^{\text{ins}}$ in MGD                   & 44.7  \\
         MGA w/ $D^{\text{pix}}$, $D^{\text{ins}}$, and $D^{\text{cat}}$ in MGD   & 47.9  \\
        \hline  
        \end{tabular}
      \label{tab:mgd}%
\end{table}

\begin{table}[!t]\footnotesize
    \centering
    \caption{Analysis of OSGF on detection performance.}
    \begin{tabular}{crcccc}
        \hline 
         & & mAP & AP$^S$ & AP$^M$ & AP$^L$\\ 
        \hline 
        \ding{202}& Baseline  & 22.0 &2.1 & 24.6 & 54.3 \\
        \ding{203}& Baseline w/ OSGF  & 24.7 & 3.2 & 29.9 & 54.8 \\
        \hline
        \ding{204}& MGA & 47.9 & 16.2 & 48.1 & 77.8 \\
        \ding{205}& MGA w/o OSGF & 45.6 & 12.2 & 46.2 & 76.1 \\
        \hline  
        \end{tabular}
      \label{tab:osgf-comp}%
\end{table}

\begin{table}[!t]\footnotesize
\centering
\caption{Comparison between different fusion methods for OSGF.}
\begin{tabular}{cccccc}
    \hline 
     S1 & S2 & \makecell[c]{w/o \\OSGF} &  \makecell[c]{Conv\\ fusion} & \makecell[c]{Average\\ fusion}  & \makecell[c]{Gated \\ fusion (ours)} \\  
    \hline 
    \checkmark   &  & 22.0   & 23.2   &  22.8  & 24.7 \\ 
    \checkmark   &   \checkmark  &  45.6  & 46.2 & 45.1 &  47.9 \\
    \hline 
    \end{tabular}
  \label{tab:aba_gate_merge}%
\end{table}

\begin{table}[!t]\footnotesize
  \centering
  \caption{Comparison of different OSGF in this work (\ie, OSGF (ours)) and conference version (\ie, OSGF (conference))~\cite{zhou2022multi}. The backbone of the detectors in this table is VGG16.}
  \setlength{\tabcolsep}{3pt}
   \begin{tabular}{rccccc}
    \hline
    method & detector  & \# Params (M) &  S1 & S1\&S2 \\
    \hline
    OSGF (conference)~\cite{zhou2022multi} & FCOS  & 17 & 23.4 & 46.8 \\
    OSGF (ours)   & FCOS  & 14 & 24.7 & 47.9 \\
    \hline
    OSGF (conference)~\cite{zhou2022multi}   & Faster-RCNN   & 55 & 33.8 & 46.1 \\
    OSGF (ours)   & Faster-RCNN   & 50 & 36.1 & 47.4 \\
    \hline
    \end{tabular}%
  \label{tab_modified_gate}
\end{table}

\begin{figure*}[!t]
	\centering
	\includegraphics[width=0.9\linewidth]{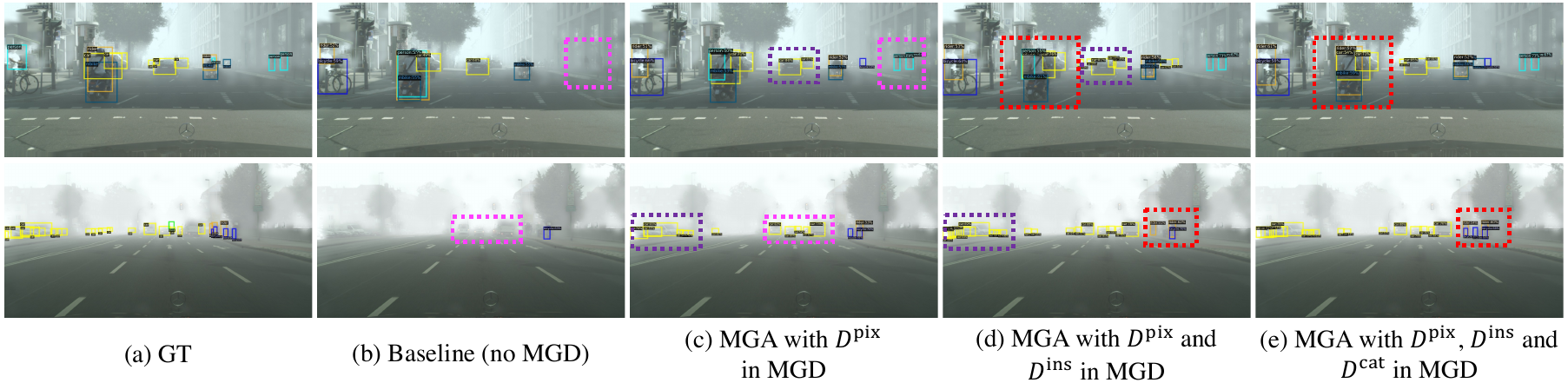}
	\caption{Qualitative visualization and comparison for the proposed modules in MGD. We can observe that, starting from the baseline (see image (b)), the usage of $D^{\text{pix}}$, $D^{\text{ins}}$, and $D^{\text{cat}}$ for feature alignment across granularities gradually improves the detection performance of our proposed MGA (see pink dashed boxes in (b) and (c), purple dashed boxes in (c) and (d), and red dashed boxes in (d) and (e) for comparison and improvements).}
	\label{fig:mgd-qual}
\end{figure*}

\begin{figure*}[!t]
	\centering
	\includegraphics[width=0.9\linewidth]{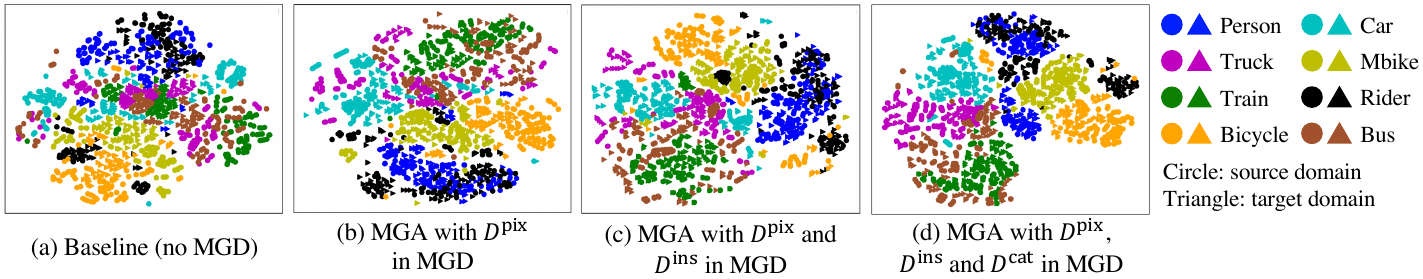}
	\caption{t-SNE visualization for the proposed modules in MGD in feature space. We can see that, from image (a) to (d), the features are gradually better aligned when using proposed alignment modules in MGD.}
	\label{fig:mgd-feat}
\end{figure*}

From Table \ref{tab:aba_dis_cls}, we obverse that the performance of detector is improved from 44.7\% to 46.2\%/46.0\%/45.2\% with 1.5\%/1.3\%/0.5\% gains by $D^\text{cen}$/$D^\text{grp}$/$D^\text{cls}$ over the baseline.  In contrast, our $D^\text{cat}$ improves the baseline detector performance from 44.7\% to 47.9\%, which shows 3.2\% performance gains over the baseline and outperforms other three category-level discriminators by $1.7\%$, $1.9\%$ and $2.7\%$, respectively, evidencing the effectiveness of our approach.

\vspace{0.3em}
\noindent
\textbf{Analysis on MGD.} To achieve domain adaption detection, we propose MGD for feature alignment by modeling dependencies across multiple granularities, including pixel, instance, and class-levels. To better understand MGD, ablation studies are carried out to analyze different modules in MGD, including pixel-level dependency ($D^{\textbf{pix}}$), instance-level dependency ($D^{\textbf{ins}}$), and category-level dependency ($D^{\textbf{cat}}$). Tab.~\ref{tab:mgd} shows the quantitative analysis for the effectiveness of modules in MGD. The baseline in Tab.~\ref{tab:mgd} indicates the original detector with OSGF but without MGD for feature alignment. From Tab.~\ref{tab:mgd}, we can see that, when applying the proposed pixel-level feature alignment ($D^{\text{pix}}$), significant improvement has been observed by increasing the mAP score of the baseline from 24.7\% to 40.6\%. When adopting the proposed instance-level feature alignment ($D^{\text{ins}}$), the detection performance is boosted from 40.6\% to 44.7\%, with a clear gain of 4.1\%. Moreover, the usage of category-level feature alignment ($D^{\text{ins}}$) further enhances the detection result by improving mAP score from 44.7\% to 47.9\% with 3.2\% performance gain. All these quantitative analyses demonstrate the effectiveness of modules in MGD. In addition, we show qualitative visualizations to exhibit the effectiveness of different feature alignment modules in MGD, as in Fig.~\ref{fig:mgd-qual}. From Fig.~\ref{fig:mgd-qual}, we can see that, the pixel-level feature alignment improves the baseline by detecting more objects (see pick dashed boxes in images (b) and (c) of Fig.~\ref{fig:mgd-qual}). The instance-level feature alignment is able to boost the performance in dealing with relatively smaller objects in crowded region (see purple dashed boxes in images (c) and (d) of Fig.~\ref{fig:mgd-qual}). The category-level feature alignment can further enhance the result by correctly detecting and classifying the objects in the target domain (see red dashed boxes in images (d) and (e) of Fig.~\ref{fig:mgd-qual}). Moreover, we also demonstrate the effectiveness of feature alignment modules in MGD in the feature space. Specifically, we display the t-SNE visualization~\cite{van2008visualizing} in feature space for the modules in MGD in Fig.~\ref{fig:mgd-feat}. From from image (a) to (d) in Fig.~\ref{fig:mgd-feat}, we can see that, the feature distributions of different categories in source and target domains are gradually better aligned when using proposed modules in MGD. 

An interesting observation from Fig.~\ref{fig:mgd-feat} is that, compared to other classes, the overlap between `person' and 'rider' seems to be more pronounced. We argue that the main reason is because the appearances of objects in `person' and `rider' categories are highly similar. In fact, the objects in `rider' are also `person', but in a different situation by riding a bicycle or a motorbike. The important cue to distinguish these two categories is the pose and surrounding bicycle or motorbike. However, due to the impact of complicated environment in domain adaption objection, it becomes more difficult to exploit these fine-grained information to distinguish them as observed in Fig.~\ref{fig:mgd-feat}. That being said, our method, compared to other approaches as shown in Tab.~\ref{tab_city_foggy}, still achieves competitive performance on `person' with the second best result and on `rider' with the best result, showing its effectiveness. We leave this as future work to explore more effective strategies to distinguish instances of similar categories such as `person` and `rider' classes.



\vspace{0.3em}
\noindent
\textbf{Effectiveness of OSGF.} To verify the effectiveness of OSGF, we conduct two sets of experiments: one is to simply apply the proposed OSGF on the baseline detector, and the other one is to remove OSGF from our MGA. To better analyze the performance, we report the results of overall mAP scores as well as AP$^L$, AP$^M$, and AP$^L$ for small, medium, and large objects, following COCO. Tab.~\ref{tab:osgf-comp} shows the experimental results. From Tab.~\ref{tab:osgf-comp}, we can observe that, when applying OSGF to the baseline (\ding{202} v.s. \ding{203}), the mAP score is improved from 22.0\% to 24.7\% with a clear gain of 2.7\%. In addition, we can also see that the AP$^{S}$, AP$^{M}$, and AP$^{L}$ for detecting small, medium, and large objects are \emph{all} increased from 2.1\% to 3.2\% (1.1\% gain), 24.6\% to 29.9\% (5.3\% gain), and 54.3\% to 54.8\% (0.5\% gain) with the help of OSGF, which shows the effectiveness of OSGF in dealing with objects of different scales for domain adaption detection. Furthermore, when removing the OSGF from MGA (\ding{204} v.s. \ding{205} in Tab.~\ref{tab:osgf-comp}), we can see that the overall mAP score is decreased from 47.9\% to 45.6\% with a drop of 2.3\%. In addition, the performance for detecting objects with different scales is degraded with AP$^{S}$ decreased from 16.2\% to 12.2\% (4.0\% drop), AP$^{M}$ from 48.1\% to 46.2\% (1.9\% drop), and AP$^{L}$ from 77.8\% to 76.1\% (1.7\% drop), which reveals the importance of OSGF in improving MGA in handing object of different scales.

\vspace{0.3em}
\noindent
\textbf{Analysis on gated fusion.} In this paper, we introduce a  gated fusion approach to improve feature representation. In order to further analyze the gated fusion, we compare it with other commonly used fusion strategies including Conv fusion and Average fusion in Table \ref{tab:aba_gate_merge} using FCOS with VGG16.  In Table \ref{tab:aba_gate_merge}, ``w/o OSGF'' means that the OSGF module is removed from our method. ``Conv fusion'' indicates the fusion weights in OSGF are obtained by using $1 \times 1$ convolution, and ``Average fusion'' means averaging features of convolution kernels in OSGF. From Table~\ref{tab:aba_gate_merge}, we can observe that, in S1, all three fusion methods can improve the performance. In specific, the OSGF with our gated fusion strategy improves the mAP score from 22.0\% to 24.7\% with 2.7\% performance gains, outperforming those using Conv and Average fusion methods with 23.2\% and 22.8\% mAP scores. Likewise, in S2, OSGF with our gated fusion achieves the best performance, evidencing the effectiveness of gated mechanism for representation learning.

In addition, from Table \ref{tab_modified_gate}, we can observe that the improved OSGF in this work contains less parameters. For example, the number of parameters are reduced from 17M to 14M for OSGF in FCOS-based MGA and from 55M to 50M in Faster-RCNN-based MGA, which is attributed to the designed shared convolutional weights. Meanwhile, we also see that better performance has been achieved using refined OSGF in this work compared that using the conference-version OSGF~\cite{zhou2022multi}, which is attributed to the overall design of our OSGF and indicates its advantages.

\begin{table}[!t]\footnotesize
\centering
\caption{Comparison between EMA and AEMA for detection performance.}
\begin{tabular}{cccccc}
    \hline
      EMA & AEMA  & mAP \\ 
    \hline
     &            &  45.3\\ 
    \checkmark &   & 46.5 \\
     & \checkmark & 47.9\\
    \hline
    \end{tabular}
\label{tab:comparison_ema_AEMA}%
\end{table}

\begin{table}[!t]
  \centering
  \footnotesize
  \caption{Comparison of parameter number, computational complexity, and running speed. The backbone of the detectors in this table is VGG16.`BT' indicates batch time (second) representing training time of the batch data.}
  \setlength{\tabcolsep}{3pt}
   \begin{tabular}{rccccc}
    \hline
    method & detector &  \# Total Params (M) & FPS & MACs & BT\\
    \hline
    Baseline  & FCOS &  87 & 23.3  & 291.5 & 0.9 \\
    CFA~\cite{DBLP:conf/eccv/HsuTLY20}  & FCOS & 177  & 17.5 & 362.1 & 1.8\\
    SCAN~\cite{DBLP:conf/aaai/Li_2022}  & FCOS & 191  & 15.2 & 389.0 & 2.3 \\
    MGA (ours)   & FCOS & 429  & 9.1 & 581.0 &  3.1 \\
    \hline
    Baseline  & Faster-RCNN & 521  & 26.1 & 263.0 & 0.7 \\
    SCL~\cite{DBLP:journals/corr/abs-1911-02559}  & Faster-RCNN & 580  & 11.8 & 347.1 & 1.3 \\
    SAPNet~\cite{DBLP:journals/corr/abs-1911-02559}  & Faster-RCNN & 556 & 25.2 & 572.5 & 1.4 \\
    MGA (ours)   & Faster-RCNN & 535 & 11.0 & 730.6 & 1.8 \\
    \hline
    \end{tabular}%
  \label{tab_complexity}
\end{table}

\vspace{0.3em}
\noindent
\textbf{Comparison of EMA and AEMA.} To show the effectiveness of our proposed AEMA, we compare it with EMA using FCOS with VGG16 as shown in Tab.~\ref{tab:comparison_ema_AEMA}. From Tab.~\ref{tab:comparison_ema_AEMA}, we obverse that EMA brings in a 1.2\% performance gain by improving the result from 45.3\% to 46.5\%. However, when using the proposed AEMA, the performance is improved to 47.9\% with 2.6\% gains, which demonstrates the superiority of our AEMA in learning better pseudo labels for detection. 

It is worth noting that, if applied on the source domain data, AEMA may encounter the overfitting issue. Specifically, as demonstrated in Tab.\ref{tab:hyperp-ab}, we can observe that, when U($a_{\delta}$, $b_{\delta}$) and U($a_{u}$, $b_{u}$) are both 1.0 and the evaluation data and the training source domain data are the same, the performance of the detector trained with AEMA method is 46.2\%, which is actually lower than that trained using EMA with 46.5\% (see Tab.~\ref{tab:comparison_ema_AEMA}) due to the overfitting issue. However, when generating the discrepancy of data distribution on the source data, this issue can be effectively alleviated. From Tab.~\ref{tab:hyperp-ab}, we can see that, when using appropriate ($a_{\delta}$, $b_{\delta}$) and ($a_{u}$, $b_{u}$), \eg, ($0.8$, $0.9$) and ($0.4$, $0.5$), the domain shift simulation effectively helps our AEMA in improving detection performance with 47.9\% mAP score, which clearly outperforms EMA with 46.5\% mAP score.

Additionally, in AEMA, we exclusively sample evaluation data from the source domain because the target domain does not provide label information for the data (note, this is determined by the definition of the unsupervised domain adaption detection) and thus the pseudo-labels generated, if using data from target domain, will contain substantial amount of noise, which is not desired for AEMA in which we need to precisely assess the performance of both student and teacher networks and reasonably integrate the two based on evaluation results. Because of this, we choose to use the source domain data only for AEMA.

\vspace{0.3em}
\noindent
\textbf{Computational complexity.} In Table \ref{tab_complexity}, we show the comparison of computational complexity between our method and other state-of-the-art approaches. As shown in Tab. \ref{tab_complexity}, From Tab.~\ref{tab_complexity}, we can observe that, all the domain adaption detection models, including CFA~\cite{DBLP:conf/eccv/HsuTLY20} and SCAN~\cite{DBLP:conf/aaai/Li_2022} based on FCOS and SCL~\cite{DBLP:journals/corr/abs-1911-02559}, SAPNet~\cite{DBLP:journals/corr/abs-1911-02559} based on Faster-RCNN, and ours, result in increased computational complexity. Compared to the state-of-the-art CFA and SCAN based on FCOS, our MGA performs better with reasonable increased complexity. In addition, on the top of Faster-RCNN, our method shows better performance and has less parameters than two recent methods SCL and SAPNet, though the computational complexity is slightly higher.

To scale with dataset size (\eg, when dealing with large-scale datasets), one possible solution is to optimize OSGF and MGD which are involved with extensive feature extraction and alignment. For example, we can adopt parallel techniques to implement the multi-stream feature extraction in OSGF. Besides, another way is to increase the training batchsize. Since the gpu memory of our method is not that large, one can accelerate learning on large-scale datsets with larger batchsize and using more gpus. Since this is beyond the goal of this work, we leave these as our future work.

\begin{figure}[!t]
	\centering
\includegraphics[width=0.8\linewidth]{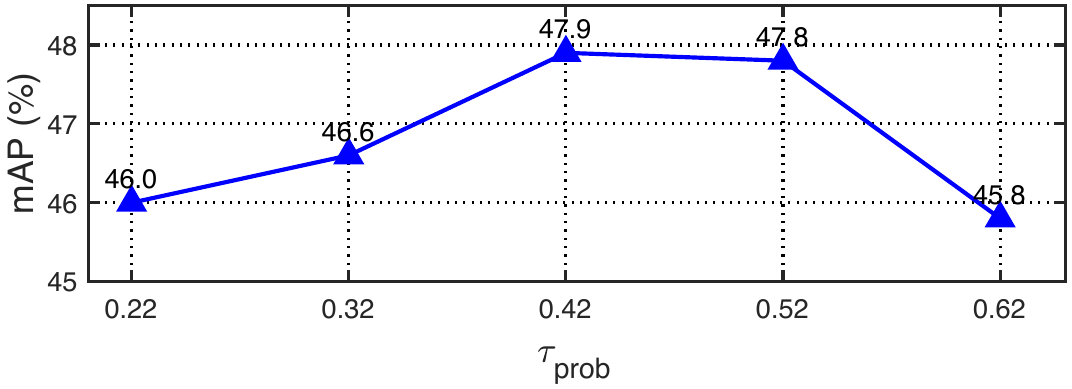}
	\caption{Effectiveness of the hyper-parameter $\tau_{\text{prob}}$.}
	\label{fig:hyper-tau}
\end{figure}

\vspace{0.3em}
\noindent
\textbf{Analysis on hyperparameter.} Similar to other popular approaches, there are a few hyperparameters in our method. In this work, the hyperparameters are empirically set based on our experimental results. To analyze these parameters and their effectiveness, we have conducted ablation experiments on crucial hyperparameters including $\tau_{\text{prob}}$ which is used to control the generation of pseudo labels in MGD and ($a_{\delta}$, $b_{\delta}$), and ($a_{u}$, $b_{u}$) in AEMA. In Fig.~\ref{fig:hyper-tau}, we show the effectiveness of $\tau_{\text{prob}}$ on the final performance. If $\tau_{\text{prob}}$ is too small, the pseudo labels may contain much noise, degrading performance. However, if $\tau_{\text{prob}}$ is too large, the number of pseudo labels may not be enough for effective learning. Therefore, we set $\tau_{\text{prob}}$ as 0.42 in our work. Tab.~\ref{tab:hyperp-ab} report the ablation analysis on ($a_{\delta}$, $b_{\delta}$), and ($a_{u}$, $b_{u}$) in AEMA. We show the experimental results with different values for ($a_{\delta}$, $b_{\delta}$), and ($a_{u}$, $b_{u}$). From Tab.~\ref{tab:hyperp-ab}, we can see that when ($a_{\delta}$, $b_{\delta}$), and ($a_{u}$, $b_{u}$) are set to (0.8, 0.9) and (0.4, 0.5), the proposed method achieves the best performance. When slightly adjusting their values, the mAP score does not change much, which shows that our approach is less sensitive to ($a_{\delta}$, $b_{\delta}$), and ($a_{u}$, $b_{u}$).

\begin{table}[!t]\footnotesize
    \centering
    \caption{Hyperparameter analysis of ($a_{\delta}$, $b_{\delta}$) and ($a_{u}$, $b_{u}$) in AEMA.}
    \begin{tabular}{ccc}
        \hline 
         ($a_{\delta}$, $b_{\delta}$) & ($a_{u}$, $b_{u}$)  & mAP \\ 
        \hline 
         (0.8, 0.9)          &    (0.8, 0.9)        &  46.9 \\ 
         (0.6, 0.7)          &    (0.8, 0.9)        &  46.4  \\ 
         (0.4, 0.5)          &    (0.8, 0.9)        &  47.6  \\ 
         
         (0.8, 0.9)          &    (0.6, 0.7)        &   47.0 \\ 
         (0.6, 0.7)          &    (0.6, 0.7)        &  47.2  \\ 
         (0.4, 0.5)          &    (0.6, 0.7)        &  46.6  \\

         (0.8, 0.9)          &    (0.4, 0.5)        &  47.9  \\ 
         (0.6, 0.7)          &    (0.4, 0.5)        &  47.8  \\ 
         (0.4, 0.5)          &    (0.4, 0.5)        &  47.0  \\
        \hline  
        U($a_{\delta}$, $b_{\delta}$)=1.0 & U($a_{u}$, $b_{u}$)=1.0  & 46.2 \\
        \hline
        \end{tabular}
      \label{tab:hyperp-ab}%
\end{table}

\section{Discussion}
\label{diss}

\subsection{Discussion on Different Scenarios}

\begin{figure}[!t]
	\centering
	\includegraphics[width=\linewidth]{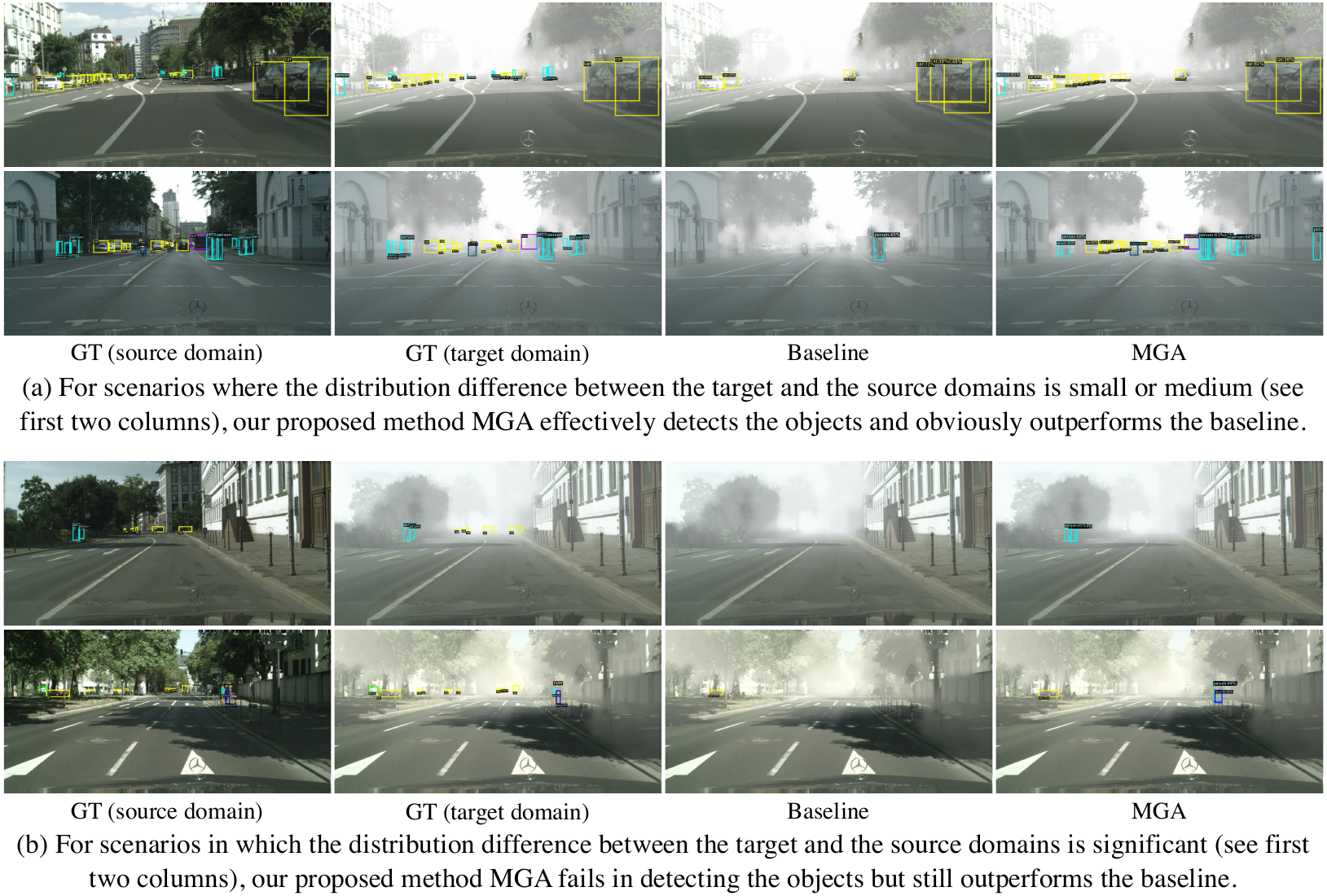}
	\caption{Illustration of MGA in different scenarios on weather adaption. Image (a) displays scenarios in which our method is effective, while image (b) demonstrates scenarios where our approach fails.}
	\label{fig:rev-fig1}
\end{figure}

Our goal is to develop a unified multi-granularity alignment framework for UDA detection. The key is to reduce differences in the feature distributions
of source and target domains by encoding dependencies across granularities including pixel-, instance-, and category-levels simultaneously for alignment. In scenarios where distribution difference between target and source domains is small or medium, our method can effectively detect objects in the target domain. Nevertheless, when the distribution difference between target and source domains are significant, our method might fail. Please note that, it is extremely difficult to quantitatively define or measure the distribution difference between source and target domains. For better illustration, we qualitatively demonstrate these scenarios with small/medium and significant distribution differences and our detection results in these situations on the setting of Cityscapes$\rightarrow$FoggyCityscapes, as shown in Fig.~\ref{fig:rev-fig1}. From Fig.~\ref{fig:rev-fig1} (a), we can see that when the fog is slight or at medium-level, out method can effectively detect the objects from the image. However, when the fog is at heavy-level, causing significant distribution difference, our method will fail, as in Fig.~\ref{fig:rev-fig1} (b). It is worth noting that, although the proposed method fails in detecting objects in scenarios with significant distribution difference between source and target domains, it still outperforms the baseline method.

\subsection{Discussion on Failure Cases}

\begin{figure}[!t]
	\centering
\includegraphics[width=\linewidth]{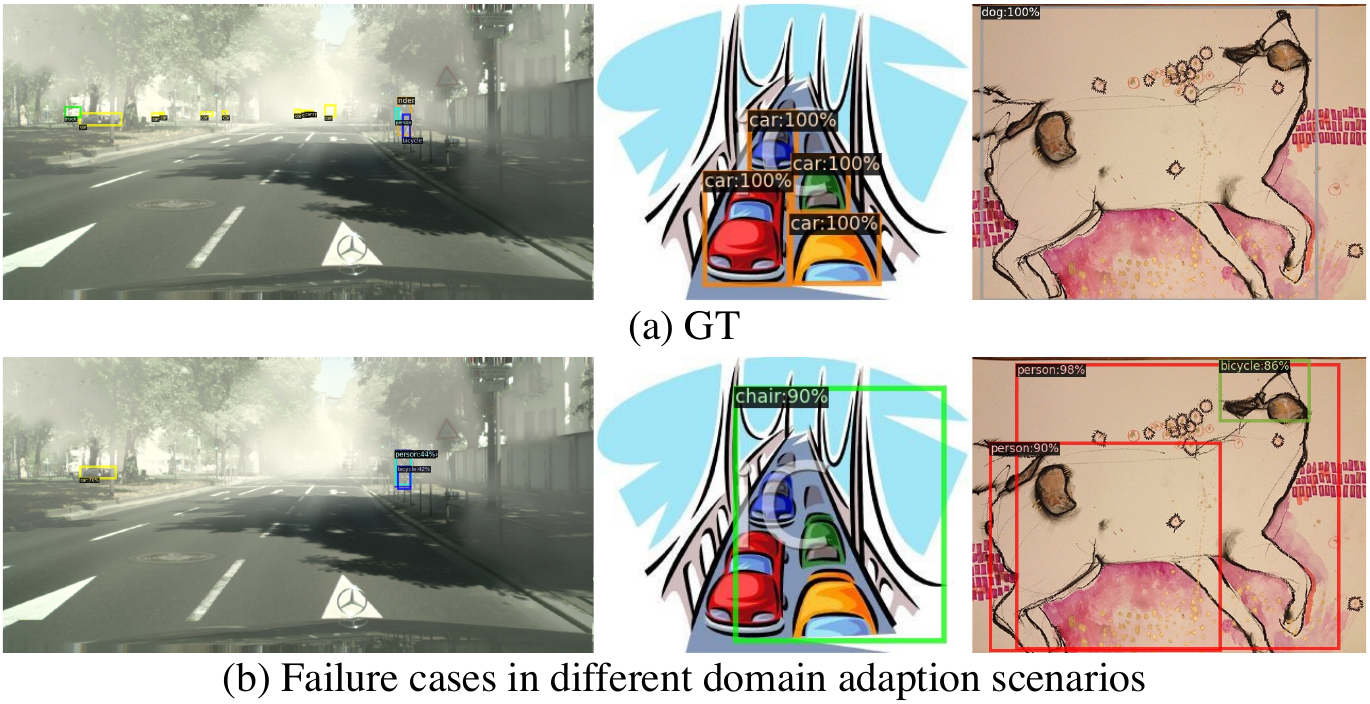}
	\caption{Demonstration of failure cases. Image (a) shows the groundtruth and image (b) the failure detection results of our method on weather adaptation from Cityscapes to FoggyCityscapes and real-to-artistic adaptation from PASCAL VOC to Clipart and Watercolor.}
	\label{fig:rev-fig3}
\end{figure}

Despite promising performance of our method, it might fail in the extremely challenging scenarios. To help readers better understand our approach, we demonstrate a few failure cases of MGA in different adaption settings, as shown in Fig.~\ref{fig:rev-fig3}. Specifically, we show the failure cases of our method on weather adaptation from Cityscapes to FoggyCityscapes and real-to-artistic adaptation from PASCAL VOC to Clipart and Watercolor. Please note that, we show failure cases in these adaption settings because there are more categories contained. From Fig.~\ref{fig:rev-fig3}, we can see that our MGA may fail in detecting objects when the distribution of source domain is too far away from that of the target domain because of the difficulty in feature alignment under these scenarios.

To better analyze the failure cases, we, from a statistical perspective, conduct experiments to show the limitation of our method under different distributions between source and target domains. Specifically, we use the domain adaption concerning degrees of difference, including \emph{small}, \emph{medium}, and \emph{large}, to measure the distribution difference between domains or the so-called difficulty. In Tab.~\ref{tab:aba_fogy_degree}, we display the comparison results of our MGA approach in weather adaptation $C\rightarrow F$ with three levels of fog density (\i.e, 0.005 for small, 0.01 for medium, and 0.02 for large), corresponding to visibility ranges of 600, 300, and 150m respectively. Note that, the definition of different levels of fog density is given by the benchmark~\cite{DBLP:journals/ijcv/SakaridisDG18}. From Tab.~\ref{tab:aba_fogy_degree}, we observe that as fog density increases, \ie, the difference of distributions between two domains increases, the performance of our MGA is degraded. This shows that when distribution of source domain is too far away from that of the target domain, it becomes difficult to effectively align features under these scenarios, degrading the proposed method. In fact, this will also degenerate other methods as shown in Tab.~\ref{tab:aba_fogy_degree}. Compared to other methods, our method still exhibits advantages under different situations.

\begin{table}[!t]\footnotesize
\centering
\caption{Performance under different levels of distribution differences, including small, medium, and large, on weather adaptation from Cityscapes to FoggyCityscapes.}
\begin{tabular}{cccc}
    \hline 
     method & Small & Medium & Large  \\ 
    \hline 
    MGA (Ours)  &  54.0   &  52.7  &  47.9 \\ 
     CFA~\cite{DBLP:conf/eccv/HsuTLY20}        &  40.5   &  39.9  &  36.0 \\ 
     SCAN~\cite{DBLP:conf/aaai/Li_2022}       &  44.7   &  42.3  &  41.5 \\ 
     SIGMA~\cite{li2022sigma}      &  46.5   &  44.9  &  43.5 \\ 
    \hline 
    \end{tabular}
  \label{tab:aba_fogy_degree}%
\end{table}

\subsection{Discussion on MGD and OSGF}

OSGF and MGD are important component of the proposed MGA. Compared to OSGF, MGD brings the major improvement in MGA for achieving domain adaption detection. This is because, for the domain adaption detection task, the main challenge lies in how to effectively align feature distributions of the source and target domains. The proposed MGD is introduced to \emph{directly solve this core challenge} in the domain adaption detection by modeling dependencies across different granularities to achieve multi-level (i.e., pixel-, instance-, and category-level) alignment, which thus significantly improves the performance of our MGA. Different from MGD directly working on aligning feature distributions, the OSGF is proposed to \emph{enhance the discriminative capacity of feature maps}. It is \emph{\textbf{not}} intended or utilized for aligning the feature distributions of the source and the target domains. Therefore, compared to the MGD module, OSGF brings relatively small improvements in our MGA. That being said, OSGF is an important component of MGA to improve the performance. As shown in Tab.~\ref{tab:osgf-comp}, OSGF improves the baseline with 2.7\% in mAP. When removing OSGF from MGA, the final performance will drop by 2.3\% in mAP, which shows that the features enhanced by OSGF can help MGD for better distribution alignment between the source and target domains and thus leads to better results.

\subsection{Future Direction}

Unsupervised domain adaption detection is an importation task in computer vision. For unsupervised domain adaption detection, one of the challenges is feature alignment between two domains under complicated scenarios. For example, the distribution of source and target domains is too far away. To deal with this, one future direction is to leverage large-scale vision-language models for improvements. These vision-language models are usually pre-trained with a large amount of data and thus have the strong capacity for zero-shot learning. We can explore this zero-shot recognition ability by extracting and injecting their features into existing domain adaption detection models to further improve performance in complex scenarios. In addition, how to distinguish target from extremely similar categories such as person and rider is another challenging problem. To handle this, one possible solution is to explore fine-grain learning to exploit more discrimination local details of the object for recognition.

\section{Conclusion}
\label{con}
In this paper, we propose a novel unified multi-granularity alignment (MGA) framework, which encodes dependencies among different pixel-, instance-, and category-levels to achieve alignment of feature distributions for unsupervised domain adaptive detection. Notably, we design the omni-scale gated fusion module with different scales and aspect rations to extract discriminative instance-level feature representation. In order to improve the quality of pseudo labels and mitigate the local misalignment problem in our MGA framework, we further propose a simple but effective dynamic adaptive exponential moving average strategy. Extensive experiments evidence the effectiveness and superiority of our MGA on different detectors for UDA detection.




\bibliographystyle{IEEEtran}
\bibliography{Robust_Domain_Adaptive_Object_Detection}


\begin{IEEEbiography}[{\includegraphics[width=1in,height=1.25in,clip,keepaspectratio]{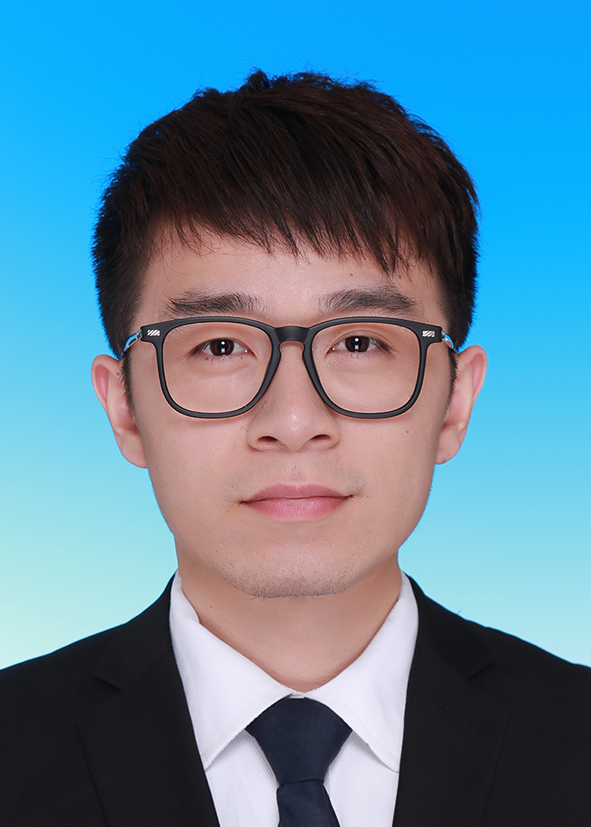}}]{Libo Zhang} received the PhD degree in computer software and theory from the University of Chinese Academy of Sciences, Beijing, China, in 2017. He is currently an Associate Research Professor with the Institute of Software, Chinese Academy of Sciences, Beijing. He has been selected as a member of the Youth Innovation Promotion Association, the Chinese Academy of Sciences, and the Outstanding Youth Scientist of the Institute of Software, Chinese Academy of Sciences. His current research interests include image processing and pattern recognition.
\end{IEEEbiography}

\begin{IEEEbiography}[{\includegraphics[width=1in,height=1.25in,clip,keepaspectratio]{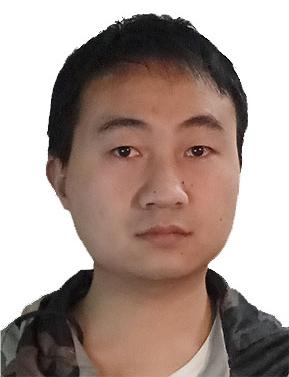}}]{Wenzhang Zhou}
received the Master’s degree from the University of Electronic Science and Technology of China in 2018. He is currently pursuing the
PhD degree with the School of Computer Science and Technology, University of Chinese Academy of Sciences, China. His research interests include visual tracking and object detection.
\end{IEEEbiography}

\vfill 

\begin{IEEEbiography}[{\includegraphics[width=1in,height=1.25in,clip,keepaspectratio]{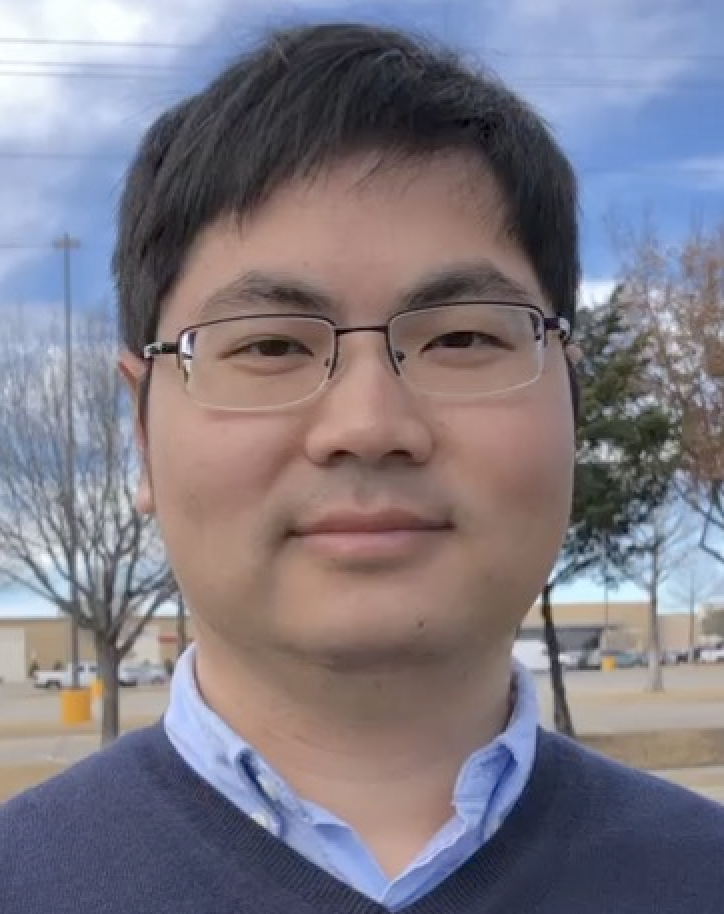}}]{Heng Fan}
received the BS degree from Huazhong Agricultural University, Wuhan, China, in 2013, and PhD degree from the Stony Brook University, Stony Brook, New York, in 2021, respectively. Currently, he is an assistant professor with the Department of Computer Science and Engineering, University of North Texas, Denton, TX USA. He has served as an Area Chair for WACV 2022, 2023 and 2024. His research interests include computer vision, machine learning, and robotic vision.
\end{IEEEbiography}

\vfill

\begin{IEEEbiography}[{\includegraphics[width=1in,height=1.25in,clip,keepaspectratio]{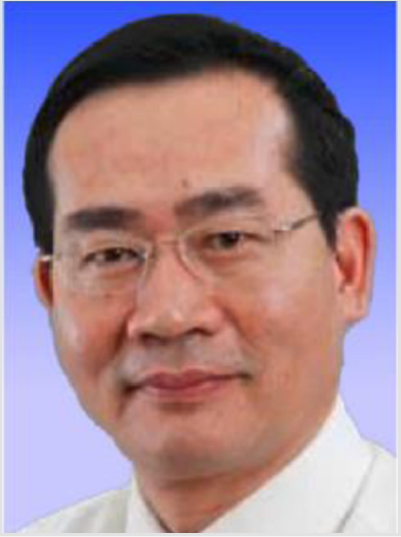}}]{Tiejian Luo} received the PhD degree in computer software and theory from the Graduate University of Chinese Academy of Sciences (UCAS), Beijing, China, in 2001. He is currently a Full Professor with the School of Computer Science and Technology, UCAS, and a Research Professor
with the Institute of Software, Chinese Academy of Sciences. He is the Director of Information Dynamics and Engineering Application Laboratory, UCAS. His current research interests include web mining and deep learning.
\end{IEEEbiography}

\vfill

\begin{IEEEbiography}[{\includegraphics[width=1in,height=1.25in,clip,keepaspectratio]{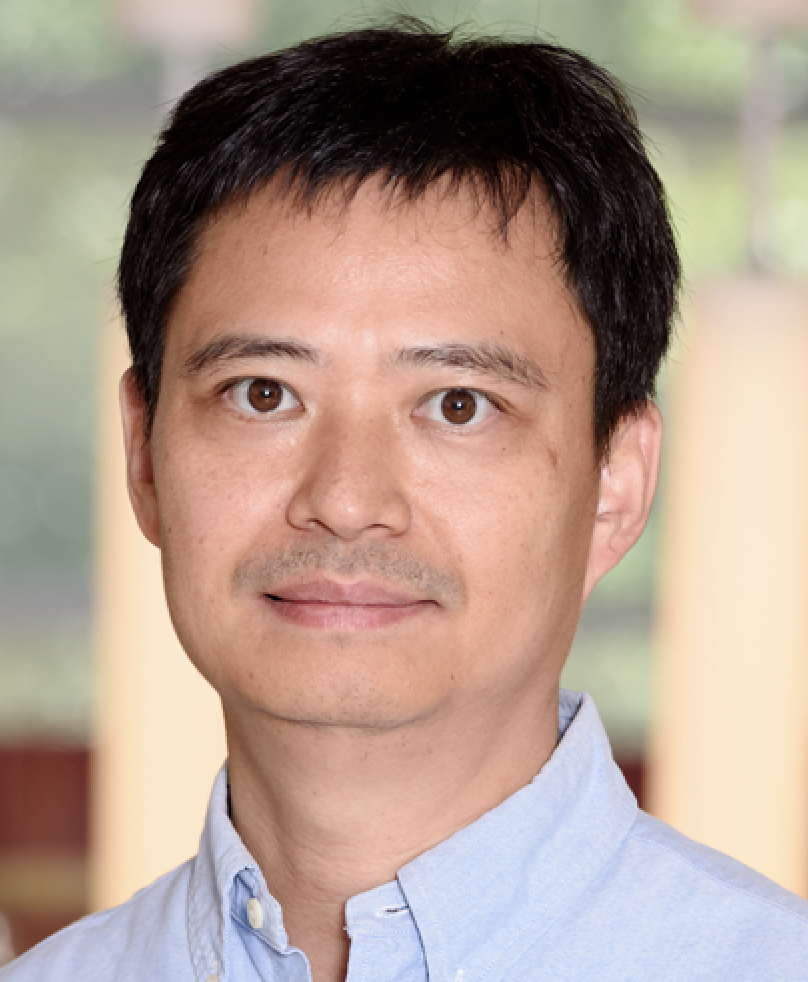}}]{Haibin Ling}
received the PhD degree from the University of Maryland, College Park, Maryland, in 2006. From 2000 to 2001, he was an
assistant researcher with Microsoft Research Asia. From 2006 to 2007, he worked as a postdoctoral scientist with the University of California Los Angeles. In 2007, he joined Siemens Corporate Research as a research scientist; then, from
2008 to 2019, he worked as a faculty member of the Department of Computer and Information Sciences, Temple University. In fall 2019, he joined Stony Brook University as a SUNY Empire Innovation Professor with the Department of Computer Science. His research interests include computer vision, augmented reality, medical image analysis, and human computer interaction. He received the NSF CAREER Award in 2014. He serves or served as an associate editor for several journals including \emph{IEEE T-PAMI}, \emph{IEEE TVCG}, \emph{PR}, and \emph{CVIU} as well as area chairs various times for CVPR, ICCV, ECCV, and WACV.
\end{IEEEbiography}


\vfill

\end{document}